\documentclass[letterpaper, 10 pt, conference]{ieeeconf}

\IEEEoverridecommandlockouts
\usepackage{iitem}
\usepackage{amsmath,amssymb}
\usepackage{mathtools}
\newcommand{\pose}[2]{\prescript{#1}{}{\mathbf{T}}_{#2}}   
\usepackage{graphicx}
\usepackage{booktabs}
\usepackage{siunitx}
\DeclareSIUnit\frames{frames}
\usepackage{multirow}
\usepackage{capt-of}
\usepackage{xcolor}
\usepackage{tikz}
\usepackage{pgfplots}
\usepackage{balance}
\pgfplotsset{compat=1.18}
\usepackage[nocompress]{cite}
\usepackage[hidelinks]{hyperref}
\hypersetup{pdftitle={Workhorse: Learning Robust Whole-Body Humanoid Loco-Manipulation from Human Data}, pdfauthor={Songbo Hu, Qiayuan Liao, Yufeng Chi, Kevin Zakka, Yakun Sophia Shao, Pieter Abbeel, Koushil Sreenath}}

\usetikzlibrary{arrows.meta,calc}
\pgfdeclarelayer{background}\pgfsetlayers{background,main}  
\graphicspath{{figures/}}
\renewcommand{\thesubsection}{\mbox{\thesection-\Alph{subsection}}}

\newlength{\gut}
\newlength{\hair}
\newcommand{\panelw}[2]{\dimexpr(#2-\numexpr#1-1\relax\gut)/#1\relax}
\newcommand{\framew}[2]{\dimexpr(#2-\numexpr#1-1\relax\hair)/#1\relax}
\newcommand{\ptwo}{\panelw{2}{\textwidth}}    
\newcommand{\pthree}{\panelw{3}{\linewidth}}  
\newcommand{\pquarter}{\framew{4}{\linewidth}}   
\newcommand{\wideh}[1]{\dimexpr(#1)*9/16\relax}
\newlength{\hwstackw}\newlength{\hwrigw}\newlength{\hwStackH}   
\newsavebox{\hwStack}

\definecolor{cHuman}{HTML}{D55E00}   
\definecolor{cIntent}{HTML}{0072B2}  
\definecolor{cRobot}{HTML}{009E73}   
\definecolor{cSynth}{HTML}{E69F00}   
\definecolor{cGray}{HTML}{8C939C}    
\definecolor{cLight}{HTML}{F3F4F6}
\definecolor{cAux}{HTML}{CC79A7}     
\newcommand{\swatch}[1]{\textcolor{#1}{\rule[-0.15ex]{1.4ex}{1.4ex}}}   

\tikzset{
  blk/.style={draw, rounded corners=2pt, align=center, inner sep=3pt,
              minimum height=0.7cm, fill=white, font=\scriptsize},
  gblk/.style={blk, draw=cGray, thick},
  hblk/.style={gblk}, pblk/.style={gblk}, tblk/.style={gblk}, sblk/.style={gblk},
  arr/.style={-{Stealth[length=2mm]}, semithick},
  lbl/.style={font=\scriptsize, align=center},
  tag/.style={font=\sffamily\scriptsize, text=white, fill=black, fill opacity=0.6,
              text opacity=1, inner sep=1.5pt, rounded corners=1pt},
}
\newlength{\figInset}   
\newlength{\figGap}     
\newlength{\figChan}    
\newlength{\figTagGap}  
\newlength{\figTagIn}   
\newlength{\figArrClear}
\newlength{\figLead}    
\newlength{\figThumbW}  
\newlength{\figRowH}    

\newlength{\figGlyphS}
\newcommand{\figGlyphW}[1]{\the\dimexpr(#1)*88/125\relax}  
\newcommand{\figGlyphFoot}[1]{\the\dimexpr(#1)*9/125\relax}

\newcommand{\panelcore}[4]{%
  \begin{tikzpicture}[inner sep=0, outer sep=0]
    \node[anchor=south west, #1] (im) {#2};
    \ifx\relax#3\relax\else
      \node[anchor=north west, tag] at ([shift={(\figTagIn,-\figTagIn)}]im.north west) {#3};\fi
    \ifx\relax#4\relax\else
      \node[anchor=north east, tag] at ([shift={(-\figTagIn,-\figTagIn)}]im.north east) {#4};\fi
  \end{tikzpicture}}
\newcommand{\panel}[4][]{\panelcore{}{\includegraphics[width=#3]{#2}}{#1}{#4}}

\newcommand{\panelhead}[2]{{%
  \tikzset{tag/.append style={font=\sffamily\footnotesize, inner sep=2.5pt}}%
  \panelcore{}{#2}{#1}{}}}
\newlength{\coverW}
\newcommand{\coverpanel}[3]{%
  \sbox0{\includegraphics{#1}}\pgfmathsetlength{\coverW}{max(#2,#3*(\wd0/\ht0))}%
  \begin{tikzpicture}
    \clip (0,0) rectangle (#2,#3);
    \node[anchor=center] at ($(0,0)!0.5!(#2,#3)$) {\includegraphics[width=\coverW]{#1}};
  \end{tikzpicture}}
\newlength{\fitH}
\newcommand{\fitpicture}[2]{%
  \setlength{\fitH}{#2}\sbox0{#1{\fitH}}\setlength{\fitH}{\dimexpr2\fitH-\ht0-\dp0\relax}#1{\fitH}}

\pgfkeys{/kicktease/.cd,
  stem/.initial = {teaser-sorting},
  panels/.initial = {5},
  left/.initial = {0.0588},
  right/.initial = {0.3529},
  bottom/.initial = {0.0923},
  top/.initial = {0.5846},
}

\newlength{\calloutW}
\tikzset{
  callout/.style={draw=cAux, line width=\calloutW},
  plate image/.style={anchor=south west},
  panel image/.style={anchor=north west},
  note/.style={font=\sffamily\scriptsize, text=cAux, inner sep=1pt, align=left, anchor=center},
  measure/.style={callout, line width=0.5\calloutW, {Bar[width=1.6mm]}-{Bar[width=1.6mm]}},
  leader/.style={callout, line width=0.5\calloutW, -{Circle[length=1.2mm]}},
  flight/.style={callout, line width=0.5\calloutW, dashed, -{Stealth[length=1.6mm]}},
  in frame/.style={shift={(#1.south west)}, x={($(#1.south east)-(#1.south west)$)}, y={($(#1.north west)-(#1.south west)$)}},
}
\newcommand{\kick}[1]{\pgfkeysvalueof{/kicktease/#1}}
\newcommand{\innerframe}[2]{%
  \begin{scope}
    \clip (#1) rectangle (#2);
    \draw[callout, line width=2\calloutW] (#1) rectangle (#2);
  \end{scope}}
\newcommand{\kicktease}[1]{%
  \begin{tikzpicture}[inner sep=0, outer sep=0]
    \node[plate image] (plate) {\includegraphics[width=#1]{\kick{stem}.jpg}};
    \begin{scope}[x={(plate.south east)}, y={(plate.north west)}]
      \coordinate (zoom sw) at (\kick{left},  \kick{bottom});
      \coordinate (zoom se) at (\kick{right}, \kick{bottom});
      \coordinate (zoom ne) at (\kick{right}, \kick{top});
      \draw[callout] (zoom sw) rectangle (zoom ne);
    \end{scope}
    \node[panel image] (panel 1) at (plate.south west)
      {\includegraphics[width=\framew{\kick{panels}}{#1}]{\kick{stem}-k1.jpg}};
    \foreach \i [remember=\i as \prev (initially 1)] in {2,...,\kick{panels}}
      \node[panel image] (panel \i) at ([xshift=\hair]panel \prev.north east)
        {\includegraphics[width=\framew{\kick{panels}}{#1}]{\kick{stem}-k\i.jpg}};
    \innerframe{panel 1.north west}{panel \kick{panels}.south east}
    \begin{scope}
      \clip (plate.north west) rectangle (panel \kick{panels}.south east);
      \draw[callout] (zoom sw) -- (panel 1.north west);
      \draw[callout] (zoom se) -- (panel \kick{panels}.north east);
    \end{scope}
  \end{tikzpicture}}

\newcommand{\catchtease}[1]{%
  \begin{tikzpicture}[inner sep=0, outer sep=0]
    \node[plate image] (catch) {\includegraphics[width=#1]{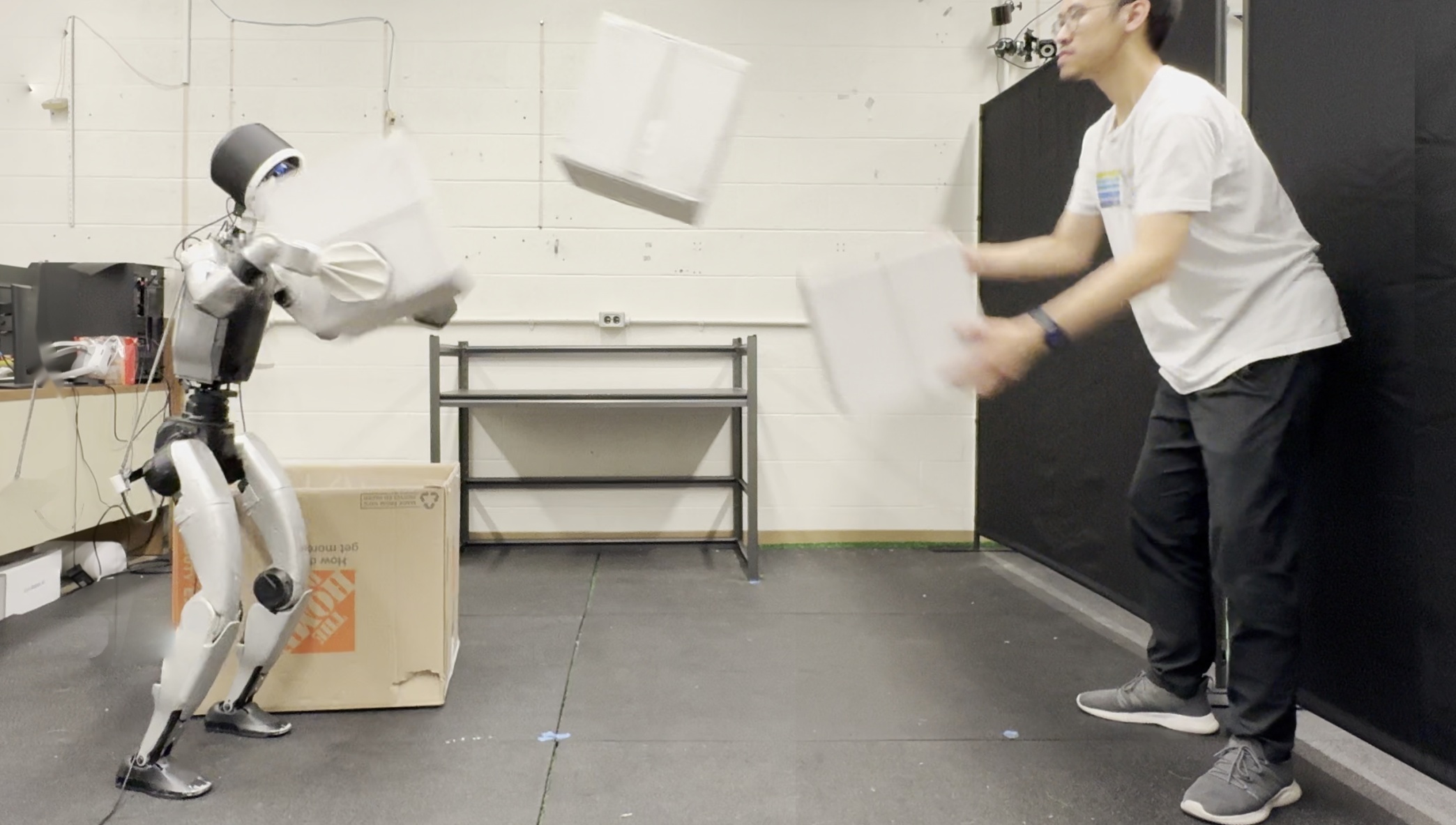}};
    \begin{scope}[in frame=catch]
      \draw[flight] plot[domain=0.61:0.26, samples=41, smooth] (\x, {-7.21*\x*\x+5.874*\x-0.314});
      \node[note, anchor=north] at (0.45,0.70) {$\sim$\qty{0.56}{\second} in flight};
    \end{scope}
  \end{tikzpicture}}

\pgfkeys{/suitcase/.cd,
  stem/.initial = {teaser-suitcase},
  panels/.initial = {8},
}

\newcommand{\suit}[1]{\pgfkeysvalueof{/suitcase/#1}}
\newcommand{\suitcasetease}[1]{%
  \begin{tikzpicture}[inner sep=0, outer sep=0, note/.append style={text=white}, callout/.append style={draw=white}]
    \node[panel image] (sp 1)
      {\includegraphics[width=\framew{\suit{panels}}{#1}]{\suit{stem}-1.jpg}};
    \foreach \i [remember=\i as \prev (initially 1)] in {2,...,\suit{panels}}
      \node[panel image] (sp \i) at ([xshift=\hair]sp \prev.north east)
        {\includegraphics[width=\framew{\suit{panels}}{#1}]{\suit{stem}-\i.jpg}};
    \begin{scope}[in frame=sp 2]
      \node[note, anchor=west] (mass) at (0.05,0.12) {\qty{14}{\kilo\gram}};
      \draw[leader] (mass.east) -- (0.58,0.28);
    \end{scope}
    \begin{scope}[in frame=sp 3]
      \draw[callout, line width=0.5\calloutW] (0.57,0.25) ellipse [x radius=0.09, y radius=0.05];
      \node[note] (brace) at (0.28,0.13) {topple};
      \draw[callout, line width=0.5\calloutW] (brace.east) -- (0.506,0.215);
    \end{scope}
    \begin{scope}[in frame=sp 5]
      \draw[measure] (0.40,0.06) -- (0.40,0.31);
      \node[note, anchor=east] at (0.38,0.19) {\qty{0.39}{\metre}};
    \end{scope}
  \end{tikzpicture}}

\pgfkeys{/perturb/.cd,
  dir/.initial = {perturb},
  panels/.initial = {8},
  rows/.initial = {6},
  tag1/.initial = {(a)},
  tag2/.initial = {(b)},
  tag3/.initial = {(c)},
  tag4/.initial = {(d)},
  tag5/.initial = {(e)},
  tag6/.initial = {(f)},
}

\newcommand{\pert}[1]{\pgfkeysvalueof{/perturb/#1}}
\newcommand{\pertframe}[2]{\includegraphics[width=\framew{\pert{panels}}{#2}]{\pert{dir}/#1.jpg}}
\newcommand{\perturbgrid}[1]{%
  \begin{tikzpicture}[inner sep=0, outer sep=0]
    \node[panel image] (pg 1-1) {\pertframe{1-1}{#1}};
    \foreach \r [remember=\r as \prow (initially 1)] in {2,...,\pert{rows}}
      \node[panel image] (pg \r-1) at ([yshift=-\gut]pg \prow-1.south west) {\pertframe{\r-1}{#1}};
    \foreach \r in {1,...,\pert{rows}} {
      \foreach \i [remember=\i as \prev (initially 1)] in {2,...,\pert{panels}}
        \node[panel image] (pg \r-\i) at ([xshift=\hair]pg \r-\prev.north east) {\pertframe{\r-\i}{#1}};
      \node[anchor=north west, tag] at ([shift={(\figTagIn,-\figTagIn)}]pg \r-1.north west) {\pert{tag\r}};
    }
  \end{tikzpicture}}

\newcommand{\clapsync}[1]{%
  \begin{tikzpicture}
    \begin{axis}[
      scale only axis, width=\linewidth, height=#1, font=\scriptsize,
      ymin=0, enlarge x limits=false, enlarge y limits=upper, clip=true,
      axis y line=none, axis x line*=bottom, tick align=center,
      ytick=\empty, xtick={0,1,2}, xticklabels={0,1,\qty{2}{\second}},
      legend style={at={(1,1)}, anchor=south east, draw=none, fill=none}, legend cell align=left,
    ]
      \addplot[cGray, dashed, ycomb, mark=none, forget plot] table[x=t_s, y expr=1, col sep=comma] {figures/clapsync-claps.csv};
      \addplot[black, semithick] table[x=t_s, y=audio, col sep=comma] {figures/clapsync.csv};
      \addlegendentry{audio onset (spectral flux)}
      \addplot[cHuman, semithick] table[x=t_s, y=clap_score, col sep=comma] {figures/clapsync.csv};
      \addlegendentry{clap score}
    \end{axis}
  \end{tikzpicture}}

\newcommand{\projname}{Workhorse}

\title{\LARGE \bf
\projname: Learning Robust Whole-Body Humanoid Loco-Manipulation from Human Data
}

\author{Songbo Hu$^{*}$, Qiayuan Liao$^{*}$, Yufeng Chi, Kevin Zakka,\\
Yakun Sophia Shao, Pieter Abbeel, and Koushil Sreenath\\
{\normalsize University of California, Berkeley}\\
{\normalsize \url{https://hybridrobotics.github.io/workhorse/}}
\thanks{$^{*}$Equal contribution.}%
}

\makeatletter
\let\@maketitlebase\@maketitle
\def\@maketitle{\@maketitlebase
\vspace{1em}
\begin{center}
  \setcounter{figure}{0}
  \begin{minipage}[b]{\ptwo}
    \centering
    \panelhead{(a) Box transport}{\kicktease{\linewidth}}
  \end{minipage}\hfill%
  \begin{minipage}[b]{\ptwo}
    \centering
    \panelhead{(b) Box catching}{\catchtease{\linewidth}}
  \end{minipage}\\[\gut]
  \begin{minipage}[b]{\textwidth}
    \centering
    \panelhead{(c) Suitcase pushing, toppling, and climbing}{\suitcasetease{\linewidth}}
  \end{minipage}
  \captionof{figure}{\textbf{\projname{} learns whole-body loco-manipulation from human data and controls a Unitree G1 with one egocentric RGB camera and proprioception.} (a)~The robot places two boxes by hand and kicks the third into the bottom shelf. The image overlays several instants, and the strip below shows the kick. (b)~The robot catches a thrown box in the air. (c)~The robot pushes the \qty{14}{\kilo\gram} suitcase to the shelf and topples the suitcase by pressing a leg against its base. It then steps up \qty{0.39}{\metre} onto the suitcase, takes a box from the top shelf, and jumps down.}
  \label{fig:teaser}
\end{center}}
\makeatother

\begin{document}

\maketitle
\thispagestyle{empty}
\pagestyle{empty}

\begin{abstract}
Humanoid robots still struggle to plan contact-rich whole-body manipulation from egocentric RGB and proprioception.
\projname{} learns such manipulation from robot-free human demonstrations.
A visual planner predicts five-link targets: the poses of the torso, both wrists, and both feet.
A reinforcement-learning whole-body tracker follows them on the robot.
Both policies train separately on the same recorded human poses, without retargeting.
We augment the training data of each policy to imitate the errors that the other makes at deployment.
On a real Unitree G1, \projname{} transport boxes with its hands and a kick, catches a thrown box, and topples and climbs a suitcase.
During box transport, we show recoveries after a person pushes the robot or takes the box away.
In a simulated copy of the demonstration room, the system completes box transport in \qty{77}{\percent} of episodes, and in \qty{64}{\percent} under \qty{40}{\newton\second} pushes.
With both policies retrained from the same demonstrations, a simulated second humanoid completes box transport in \qty{83}{\percent} of episodes without pushes.
\end{abstract}

\section{INTRODUCTION}

\begin{figure*}[t]
  \centering
  \input{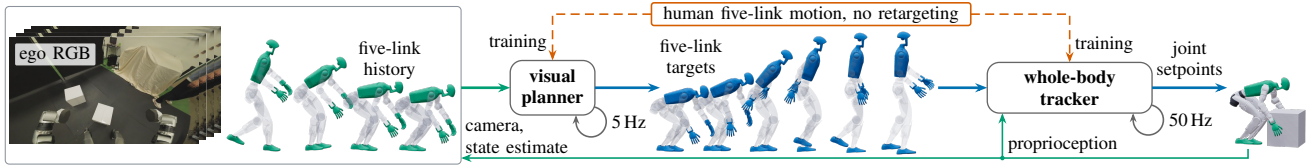}%
  \caption{Overview of \projname{}. The five links are the torso, the two wrists, and the two feet, drawn in color on a gray body shown for illustration. At deployment, the visual planner reads egocentric RGB images and the robot's five-link history, and predicts five-link targets. The whole-body tracker follows the targets and outputs joint setpoints. Dashed arrows mark training: the same human five-link motion trains both policies. Color marks the source: \swatch{cHuman}~human, \swatch{cIntent}~policy, \swatch{cRobot}~robot.}
  \label{fig:framework}
\end{figure*}

We want humanoid robots to interact with the world using whole-body motions, much like humans do.
Humans can squat to lift a box, brace a leg against an object to push or topple it, and climb onto objects to reach higher places.
Although humanoid robots can execute such whole-body motions using learned motion trackers~\cite{liao2025beyondmimic, chen2025gmt, luo2026sonic}, these methods require a whole-body reference trajectory to be specified.
For manipulation tasks, however, no such reference is available, so the robot must plan its own whole-body motion to perform the task.
Doing so from only egocentric RGB observations and proprioception is challenging, and becomes even more difficult when the robot must move dynamically, respond to changes in the scene, and recover from unexpected perturbations.

The robot can learn to plan whole-body motions for manipulation tasks from demonstrations.
Teleoperation captures demonstrations directly on the robot, but each demonstration requires both a robot and a skilled operator~\cite{he2024omnih2o,ze2025twist2}.
Robot-free rigs instead record human motion without a robot in the loop~\cite{nai2026humanoid, wang2026bifrostumi, xu2026hommi}.
For example, HuMI~\cite{nai2026humanoid} uses wearable trackers on the pelvis and feet to capture whole-body motion.
However, in most tasks studied with either approach, the feet are primarily used for locomotion rather than as active tools for manipulation.
Moreover, these tasks largely remain quasi-static, with little emphasis on dynamic motion or recovery from perturbations.

We present \projname{} (Fig.~\ref{fig:teaser}), which learns whole-body loco-manipulation from robot-free human demonstrations.
A visual planner and a whole-body tracker (Fig.~\ref{fig:framework}) communicate through a shared interface consisting of the poses of the torso, both wrists, and both feet.
Because the interface specifies task-space poses rather than robot-specific joint angles, it is naturally embodiment-agnostic.
During training, the poses come directly from human demonstrations, without requiring retargeting to a robot.

We demonstrate \projname{} on a real Unitree G1 across three whole-body loco-manipulation tasks: box transport, box catching, and suitcase pushing, toppling, and climbing.
Each task has its own planner and whole-body tracker.
For box transport, we demonstrate recovery from perturbations such as a person pushing the robot or removing the box.
In simulation, we use the same human demonstrations to train both the planner and tracker for a second humanoid, demonstrating transfer across embodiments.

Our contributions are:
\begin{itemize}
    \item \textbf{Robot-free human data collection.} We develop a system with five wearable trackers and a chest-mounted camera for recording whole-body demonstrations without a robot.
    \item \textbf{Visual planner and whole-body tracker.} We train the visual planner and whole-body tracker separately from the same five link poses extracted from the human demonstrations, without robot-specific retargeting.
    Data augmentation prepares each policy to handle errors from the other, with the tracker's reward designed for accurate wrist tracking and robust foot placement.
    \item \textbf{Closed-loop evaluation in simulation.} We reconstruct the demonstration environment as a Gaussian splat and test the complete system in closed loop.
\end{itemize}

\section{RELATED WORK}

This section reviews whole-body controllers, recording devices, and policies learned from human demonstrations, and each part ends with what \projname{} does differently.

\subsection{Humanoid Whole-Body Controllers}

Motion trackers reproduce a whole-body reference motion~\cite{liao2025beyondmimic,chen2025gmt,luo2026sonic}.
Some tolerate external disturbances~\cite{zhang2025track} or noisy reference motions~\cite{ma2026robust}.
Other controllers accept a sparse or latent command~\cite{he2024hover,tessler2024maskedmimic,li2025bfm,li2026anybody}.
The command of all these controllers is usually a retargeted motion or a teleoperator's input, which most of them treat as ground truth.
A planner's command drifts until the planner sends a new plan, and then it jumps.
We train the whole-body tracker on such commands (\S\ref{sec:tracker_aug}).

\subsection{Human Data Collection Devices}

Marker-based optical capture~\cite{mahmood2019amass} is precise but expensive.
The pose estimate of inertial sensors worn on the body drifts~\cite{yi2021transpose}.
UMI~\cite{chi2024universal} and DexUMI~\cite{xu2025dexumi} put a device with a camera in the demonstrator's hands, and each device recovers its own pose by odometry.
A whole-body version needs five such devices, including one on each foot.
Each device has its own drift and its own clock, and odometry from a camera on a foot is poorly conditioned.
An arm exoskeleton with a back-mounted LiDAR measures the arms and the torso~\cite{zhong2025humanoidexo}, but it constrains the demonstrator.
Our rig is lightweight and quick to set up.
It reports the poses of the torso, the wrists, and the feet together, in one gravity-aligned frame.

\subsection{Humanoid Policies from Human Demonstrations}

Teleoperation~\cite{ze2025twist2,he2024omnih2o} records the demonstration on the robot itself, so the data needs no conversion.
Its interface can limit how reactive and how dynamic the demonstration is.
Robot-free rigs record the poses of a few key body parts and RGB video~\cite{nai2026humanoid,wang2026bifrostumi,xu2026hommi,zhao2026halomi,shi2026egohumanoid}.
To our knowledge, the feet only walk in the tasks reported with these rigs.
Some of these systems turn the human motion into joint references for a whole-body controller by inverse kinematics or retargeting~\cite{nai2026humanoid,wang2026bifrostumi}.
Other systems retarget human motion capture or video and train a whole-body controller on it~\cite{fu2024humanplus,yin2025visualmimic,weng2025hdmi,zhao2025resmimic,yang2026zerowbc}.
Some of them use the feet on objects, for example to kick a box or a door~\cite{yin2025visualmimic,weng2025hdmi}.
Retargeting is specific to one robot, and it changes the demonstrated motion.
In \projname{}, the five recorded link poses are the training target of both policies, with no retargeting (\S\ref{sec:interface}).
A new humanoid needs five measurements on its body, its simulation model, and retraining of both policies from the same demonstrations (\S\ref{sec:crossemb}).

\section{HUMAN DATA COLLECTION}
\label{sec:data}

\begin{figure}[t]
  \centering
  \sbox0{\includegraphics{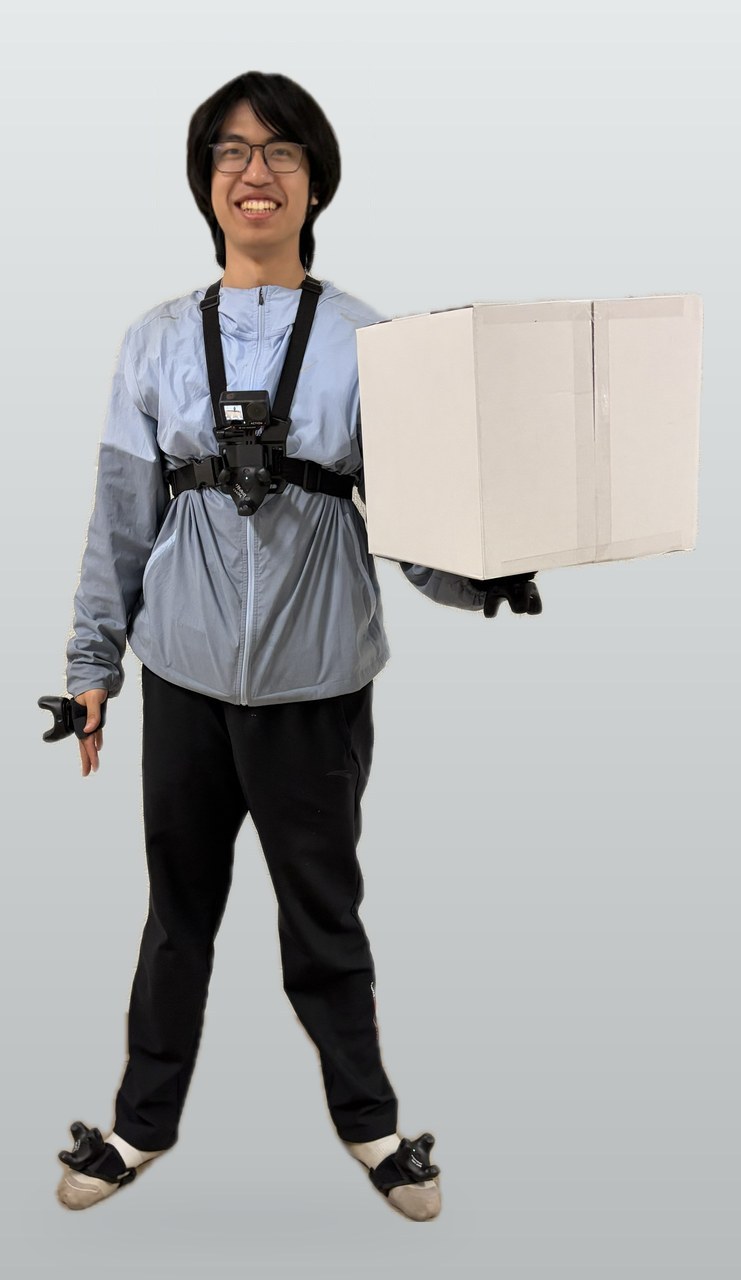}}\pgfmathsetmacro{\rigAspect}{\wd0/\ht0}%
  \sbox0{\includegraphics{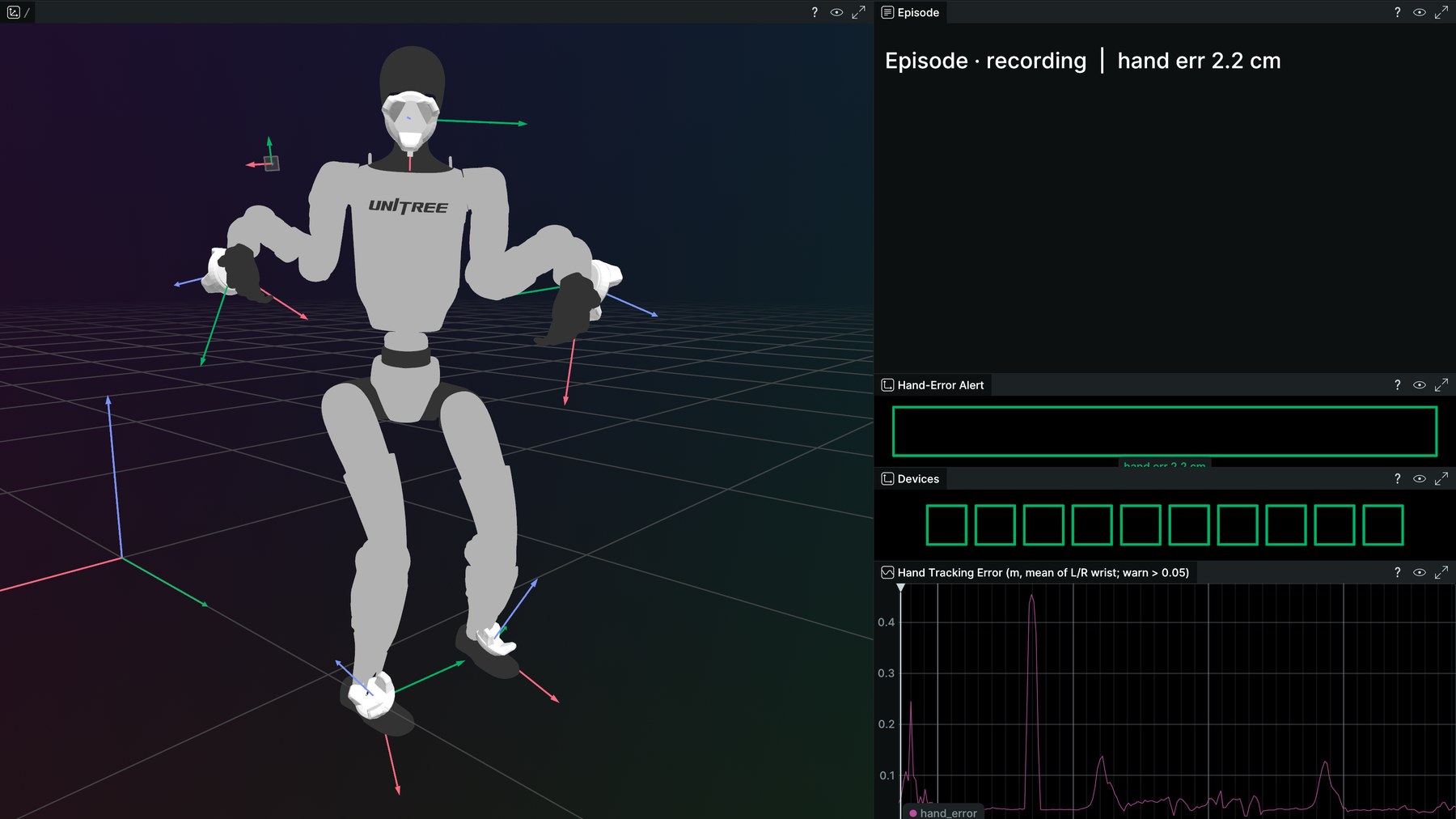}}\pgfmathsetmacro{\previewAspect}{\ht0/\wd0}%
  \pgfmathsetlength{\hwstackw}{(\linewidth-\gut*(1+\rigAspect))/(1+\rigAspect*(\previewAspect+9/16))}%
  \setlength{\hwrigw}{\dimexpr\linewidth-\hwstackw-\gut\relax}%
  \sbox{\hwStack}{\begin{minipage}[b]{\hwstackw}
    \panel[(b)]{hw_ik}{\linewidth}{}\\[\gut]
    \panelcore{}{\fitpicture{\clapsync}{\wideh{\linewidth}}}{(c)}{}
  \end{minipage}}%
  \setlength{\hwStackH}{\dimexpr\ht\hwStack+\dp\hwStack\relax}%
  \begin{minipage}[b]{\hwrigw}
    \panelcore{}{\coverpanel{rig}{\linewidth}{\hwStackH}}{(a)}{}
  \end{minipage}\hfill\usebox{\hwStack}
  \caption{Data collection. (a)~The rig: one chest camera with a wearable tracker on its mount, and one wearable tracker on each hand and each foot. (b)~The live preview: the robot posed by real-time inverse kinematics, the hand error of this solution, and the visibility of each wearable tracker. (c)~Clock synchronization on the three opening claps. The curves are the clap score of the two wearable trackers on the hands and the camera's audio onset, each scaled to unit peak. Dashed lines mark the matched claps.}
  \label{fig:hw}
\end{figure}

Both policies learn directly from the human demonstrations, so their quality limits how well the behavior transfers to the robot.
This section describes the rig (\S\ref{sec:rig}), the recording protocol (\S\ref{sec:protocol}), and the synchronization and processing (\S\ref{sec:sync}).
The protocol splits each session into repetitions, and it perturbs only the object, never the demonstrator.

\subsection{Rig}
\label{sec:rig}

The rig tracks the hands and the feet, which make the contacts, and the chest camera, which records what the demonstrator sees.
Five wearable trackers (HTC Vive) sit on the chest camera mount, the backs of the hands, and the tops of the shoes (Fig.~\ref{fig:hw}(a)).
Four Lighthouse base stations locate them at \qty{60}{\hertz} in a \mbox{$5\,\mathrm{m}\times5\,\mathrm{m}$} area, in one gravity-aligned world frame.

One DJI Osmo Action 4 on the chest records \ang{129} near-fisheye video at \qty{30}{\frames\per\second} with a rolling shutter.
It saves the video to its own SD card.
The robot carries the same camera, which streams over USB at deployment.
The planner therefore receives images with the same lens distortion and rolling shutter in training and at deployment.
We do not remove the lens distortion from the images.

\subsection{Recording Protocol}
\label{sec:protocol}

Each session is one continuous recording in which the demonstrator repeats the task.
A T-pose before each repetition marks where to split the session.

As in HuMI~\cite{nai2026humanoid}, we solve inverse kinematics on the robot model in real time during recording.
A live preview shows the result and flags a hand that is out of reach, and the demonstrator then corrects the pose (Fig.~\ref{fig:hw}(b)).
The camera is on the chest, so the demonstrator turns the torso to look for an object.

Across sessions, we vary the lighting and the object placement.
We also perturb the object during recording to widen the data coverage.
We never perturb the demonstrator, so every recorded pose is intended motion.
One demonstrator recorded all the data: \qty{2.2}{\hour} for box transport, \qty{1.2}{\hour} for the suitcase task, and \qty{13}{\minute} for box catching.

\subsection{Synchronization and Processing}
\label{sec:sync}

The camera and the wearable trackers record two streams on different clocks.
To align them, the demonstrator claps three times at the start of each session.
We compute a clap score from the distance between the two wearable trackers on the hands and from their velocity.
We match its three peaks to the three clap onsets in the camera audio, which gives the clock offset (Fig.~\ref{fig:hw}(c)).
The match minimizes the difference between the clap intervals of the two streams.
After synchronization, the two streams disagree by \qty{5.5}{\milli\second}.
We resample the wearable-tracker poses from \qty{60}{\hertz} to \qty{50}{\hertz}, the rate of the whole-body tracker.
We interpolate the samples that a wearable tracker misses when it briefly loses sight of the base stations.
Images stay at \qty{30}{\frames\per\second}, and we select them by timestamp.

After processing, each session holds the five wearable-tracker poses at \qty{50}{\hertz} in one world frame and the video on the same clock.

\section{HUMANOID SKILL TRANSFER}
\label{sec:transfer}

This section turns the human demonstrations into two policies and runs them on the robot (Fig.~\ref{fig:framework}).
Both policies share one interface, the recorded poses of five links (\S\ref{sec:interface}).
The visual planner trains on images edited to show the robot and on histories with added drift (\S\ref{sec:planner}).
The whole-body tracker trains on commands that drift and jump as they do at deployment (\S\ref{sec:tracker}).
At deployment, timestamps align the two policies, which run at different rates and with different delays (\S\ref{sec:deploy}).

\subsection{Human Motion as Interface}
\label{sec:interface}

The interface is the poses of five links: the torso, the two wrists, and the two feet.
It contains no joint angles, and it is the recorded human motion itself.
If the whole-body tracker follows it exactly, the robot reproduces the demonstrated motion, and its camera follows the demonstrator's camera trajectory.

\begin{figure}[t]
  \centering
  {\scriptsize\makebox[\pquarter]{walk}\kern\hair%
   \makebox[\pquarter]{squat}\kern\hair%
   \makebox[\pquarter]{lift}\kern\hair%
   \makebox[\pquarter]{carry}}\\[\gut]
  \panel[(a)]{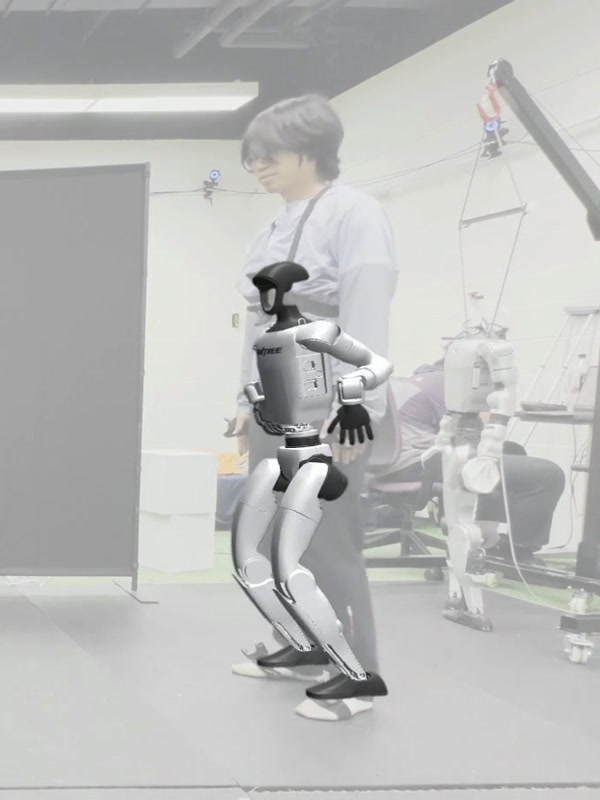}{\pquarter}{}\kern\hair%
  \panel{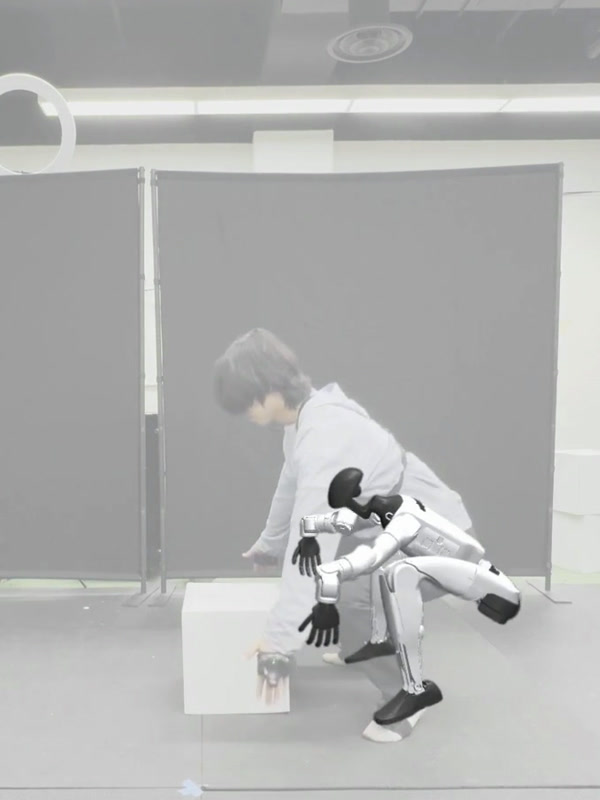}{\pquarter}{}\kern\hair%
  \panel{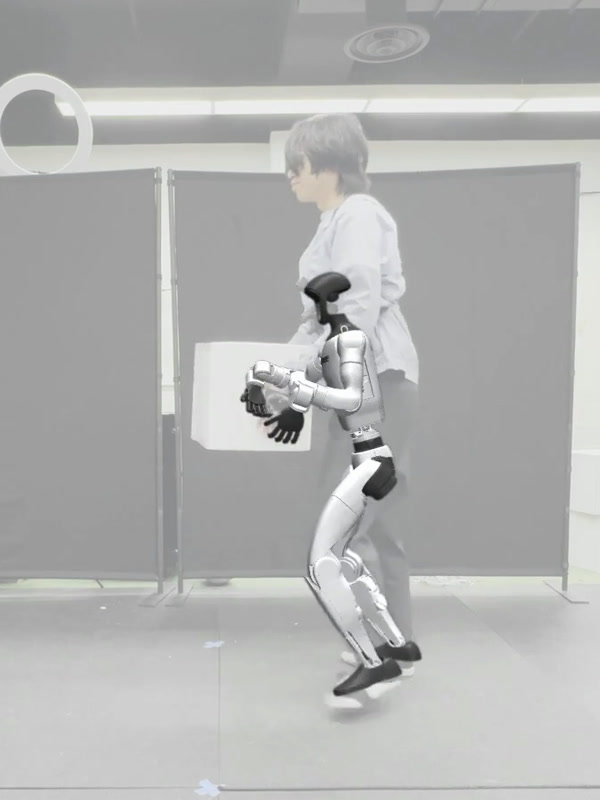}{\pquarter}{}\kern\hair%
  \panel{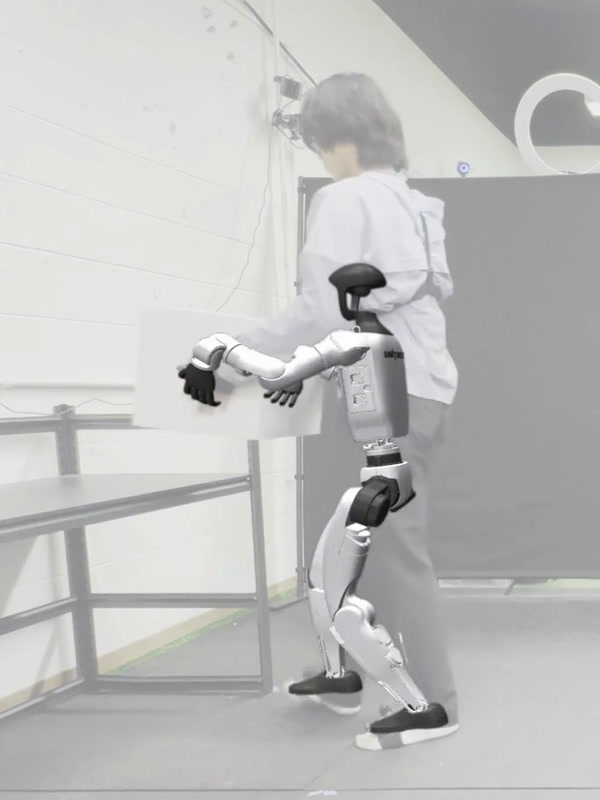}{\pquarter}{}\\[\gut]
  \panel[(b)]{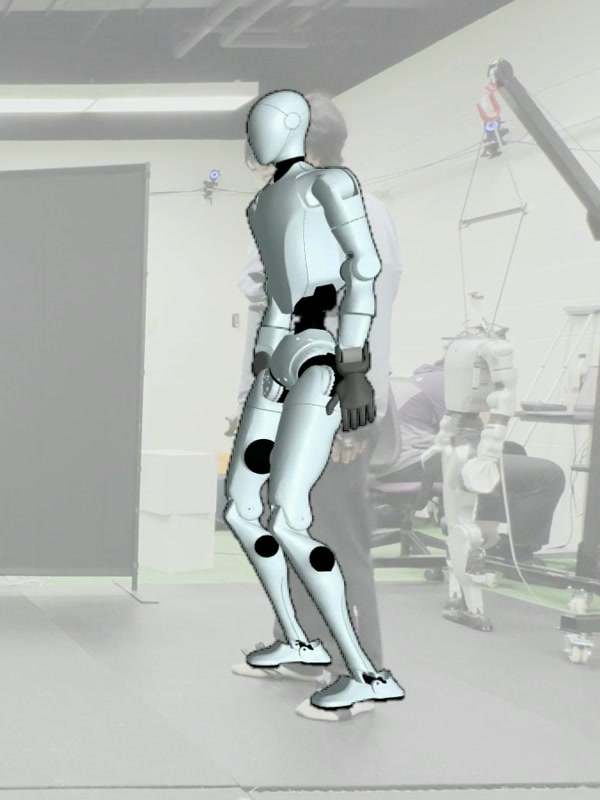}{\pquarter}{}\kern\hair%
  \panel{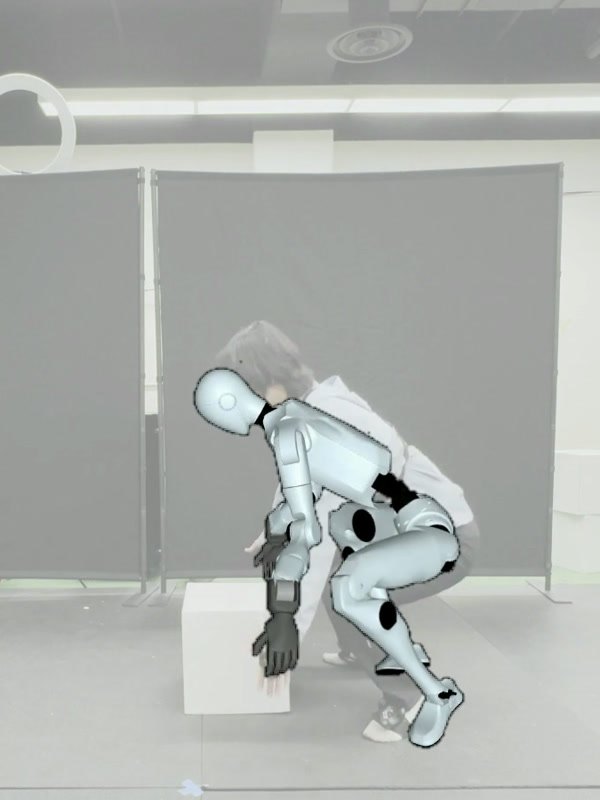}{\pquarter}{}\kern\hair%
  \panel{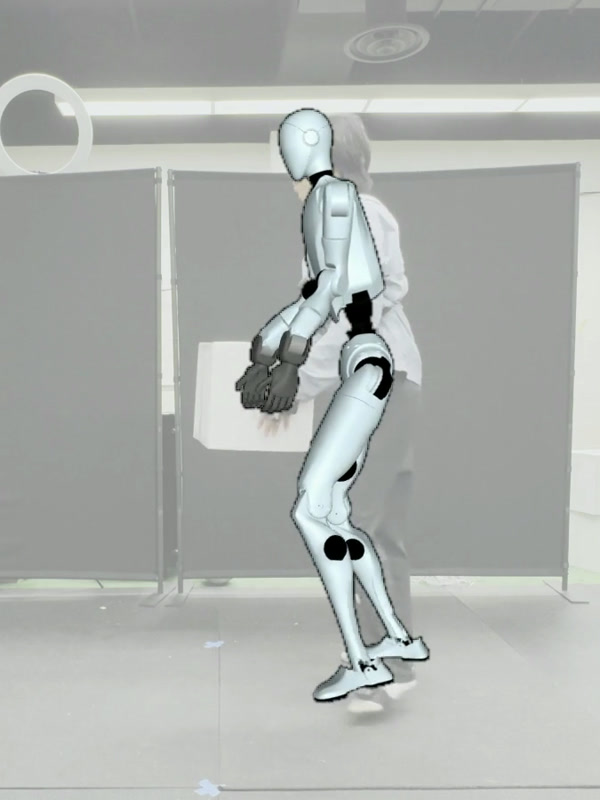}{\pquarter}{}\kern\hair%
  \panel{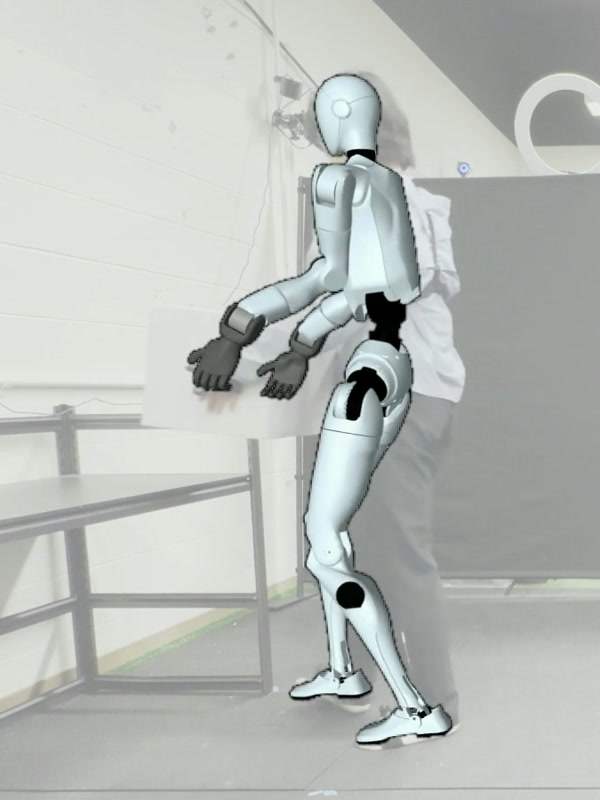}{\pquarter}{}
  \caption{One interface, two embodiments. For illustration, we pose each robot by inverse kinematics on the same demonstration and draw it over the demonstrator. (a)~Unitree G1, and (b)~Unitree H2.}
  \label{fig:bodies}
\end{figure}

\subsubsection{From Human Motion to Five-Link Targets}
\label{sec:align}

We index the five links by $b$.
Throughout, $\pose{X}{Y}$ is the pose of frame $Y$ expressed in frame $X$.
The wearable tracker for link $b$ reports the pose $\pose{W}{K_b}(t)$ of its own frame $K_b$ in a fixed world frame $W$.
For each link we choose an alignment frame $A_b$.
It is at the camera for the torso, at the palm center for each wrist, and at the sole center for each foot.
We measure two constant poses of $A_b$: $\pose{K_b}{A_b}$ in the wearable-tracker frame $K_b$, and $\pose{L_b}{A_b}$ in the robot's link frame $L_b$.
We call $\pose{L_b}{A_b}$ the mount of link $b$, because it fixes where the alignment frame sits on the robot's link.
The robot's alignment frame has to reach the pose that the demonstrator's alignment frame had, which gives the link target
\begin{equation}
  \pose{W}{L_b}(t)=\pose{W}{K_b}(t)\,\pose{K_b}{A_b}\,
  \bigl(\pose{L_b}{A_b}\bigr)^{-1}.
  \label{eq:align}
\end{equation}
The target is therefore the recorded pose moved by a fixed rigid offset, with no retargeting.
The planner predicts these five poses as the five-link targets, and the whole-body tracker receives them as the five-link command.

We never scale the human motion, so the robot has to reach the exact recorded poses.
A robot whose body is closer to the human body reaches them more easily (Fig.~\ref{fig:bodies}).
The five mounts are the only robot-specific quantities of the interface.
For a second humanoid, we measure these mounts and replace the robot model in the simulator and in the renderer.
We then retrain both policies from the same demonstrations (\S\ref{sec:crossemb}).
The tracker controls a new body, and the planner sees this body in its images and predicts five-link targets that depend on the mounts.

\subsubsection{Torso Frames}
\label{sec:bridge}

Both policies use the five link poses in a torso frame, so the poses do not depend on where the robot stands or faces.
The two policies use different torso frames.
The planner predicts a chunk, a sequence of five-link targets that starts at its observation time $t_0$, in the torso frame of $t_0$.
The whole-body tracker needs the next \qty{0.2}{\second} of this chunk in the torso frame of its control step $t$.
To change the frame from $t_0$ to $t$, we only need the torso's motion between the two times.
The onboard state estimator gives this motion, so the robot needs no external tracking.
The estimator is a Kalman filter that fuses the contact kinematics of the legs with the measured acceleration.
The planner replans at \qty{5}{\hertz}, so $t-t_0$ stays short, but the estimate still drifts.
We train the whole-body tracker on this drift (\S\ref{sec:tracker_aug}).

\subsection{Visual Planner}
\label{sec:planner}

The planner maps egocentric images and a five-link history to five-link targets.
On the robot, its input differs from the demonstrations in two ways.
The recorded images show a person, but the robot's camera sees the robot's own body.
The recorded history is clean, but the history on the robot is drifted: it contains the tracking error of the whole-body tracker and the effect of pushes.
The second difference grows in closed loop: the tracker follows a chunk predicted from a drifted history less well, and the next history drifts more.
We therefore edit the recorded images so that they show the robot (\S\ref{sec:planner_pixels}), and we add drift to the recorded histories (\S\ref{sec:planner_drift}).

\subsubsection{Observation, Output, and Architecture}
\label{sec:planner_io}
The planner reads an image history and the five-link history, which both end at $t_0$, the capture time of the newest image.
The image history has four images, \qty{0.24}{\second} apart.
The five-link history has nine samples of the five link poses, \qty{0.12}{\second} apart, in the torso frame of $t_0$.
Each pose, in the history and in the chunk, is a position and a 6D rotation, the first two columns of the rotation matrix.
Poses in the torso frame do not show how the torso tilts or how high it is.
Each sample of the five-link history therefore also includes the gravity direction in its own torso frame, which we call projected gravity.
Each sample also includes the torso height in the world frame.
The planner outputs a chunk of 58 actions, and each action holds the five-link targets of one time step.
The actions are $\Delta t=\qty{0.02}{\second}$ apart, so the chunk spans \qty{1.16}{\second}.

The planner is a flow-matching policy~\cite{lipman2022flow,liu2022flow} with a diffusion transformer backbone~\cite{peebles2022scalable}.
A ResNet-18 trained from scratch encodes each image into one token, and each sample of the five-link history gives one token.
The backbone reads these tokens through cross-attention.
Inference takes 10 Euler steps.

\begin{figure}[t]
  \centering
  \panel[(a)]{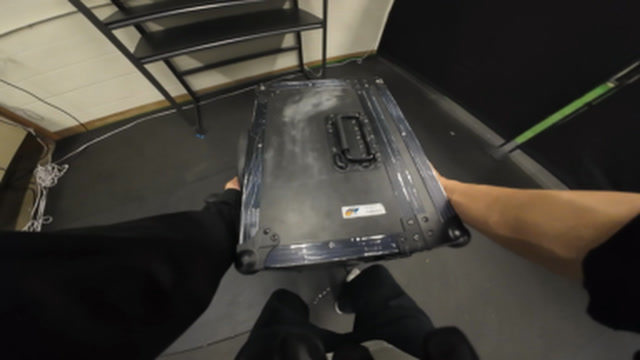}{\pthree}{raw}\hfill%
  \panel{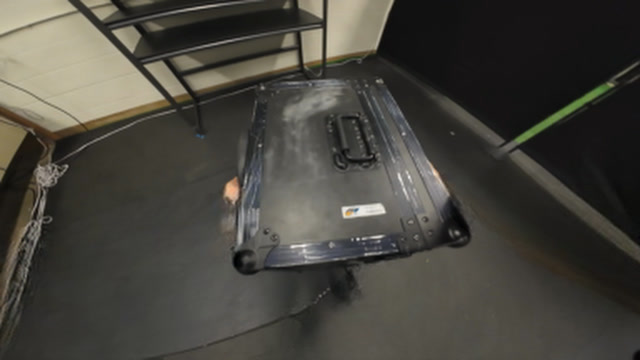}{\pthree}{person out}\hfill%
  \panel{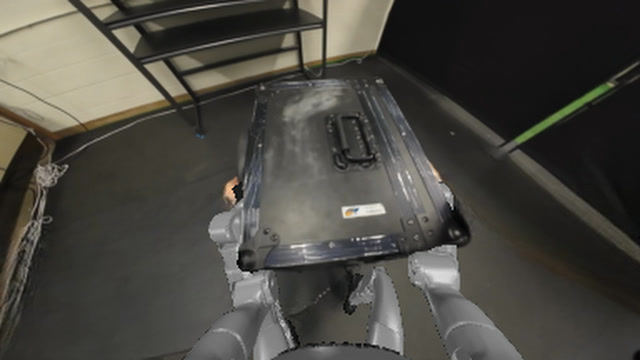}{\pthree}{robot in}\\[\gut]
  \panel[(b)]{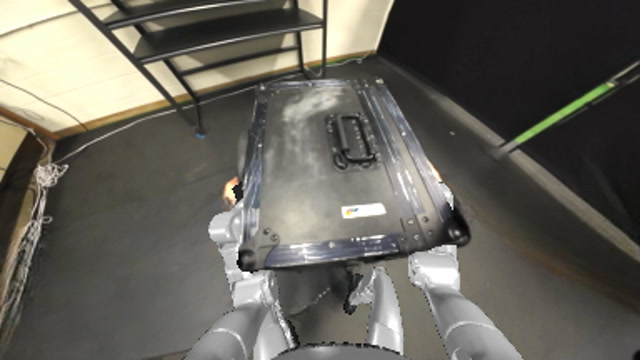}{\pthree}{brightness}\hfill%
  \panel{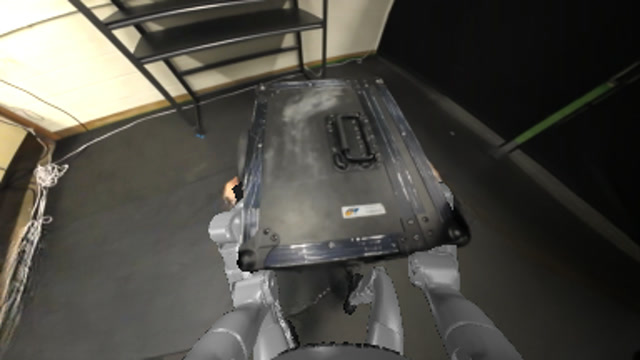}{\pthree}{lighting gradient}\hfill%
  \panel{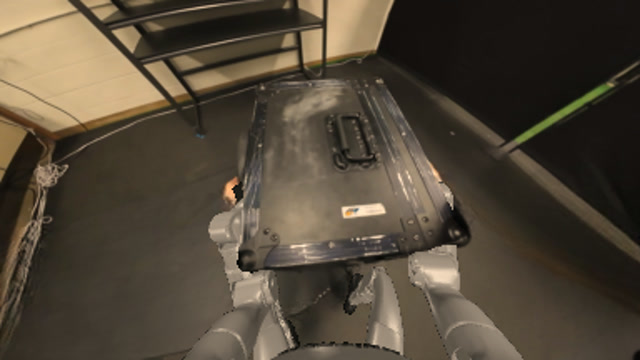}{\pthree}{contrast}\\[\gut]
  \panel[(c)]{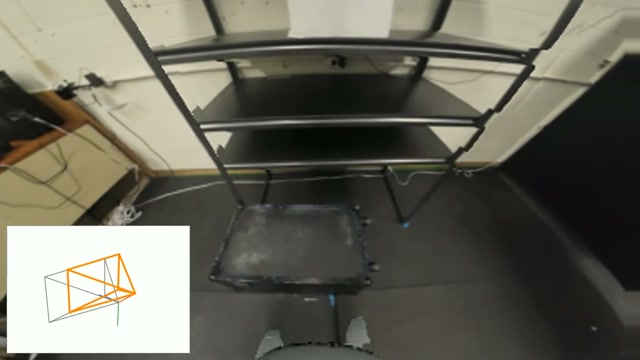}{\pthree}{}\hfill%
  \panel{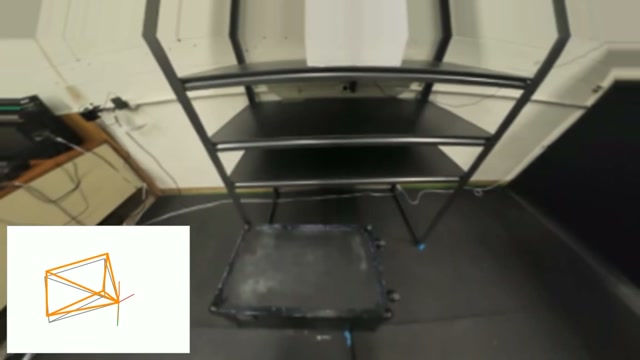}{\pthree}{}\hfill%
  \panel{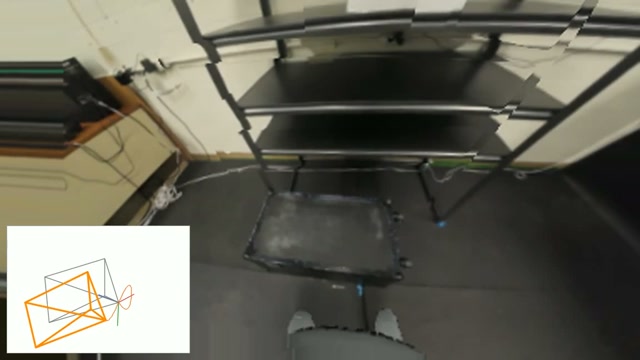}{\pthree}{}
  \caption{Image editing and augmentation for the planner. (a)~Person-to-robot edit. (b)~Photometric augmentation: brightness, lighting gradient, and contrast. (c)~Image warp of the drift augmentation for three camera displacements. Each inset shows the camera before (gray) and after (orange) the displacement.}
  \label{fig:imgaug}
\end{figure}

\subsubsection{Person-to-Robot Image Edit}
\label{sec:planner_pixels}
We edit the recorded images once, offline (Fig.~\ref{fig:imgaug}(a)).
SAM~3~\cite{carion2025sam} segments the person and the carried object, and LaMa~\cite{suvorov2021resolution} inpaints the person's pixels.
MoGe-2~\cite{wang2025moge} estimates the metric depth of the inpainted background, and we calibrate its scale against the floor plane.
MuJoCo renders the robot in the IK pose of~\S\ref{sec:tracker_reward}, seen from the recorded camera pose.
We paint the robot over every pixel whose depth lies behind it.

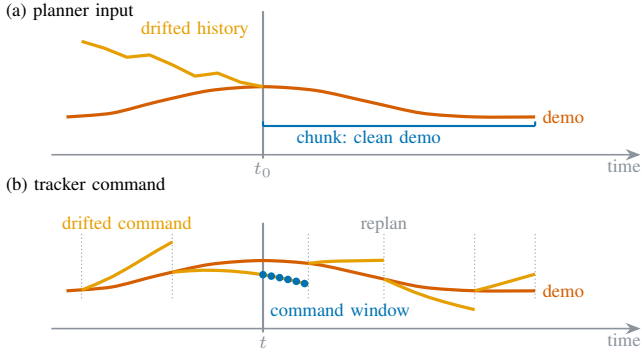
\begin{figure}[t]
  \centering
\newcommand{\nowFrac}{0.3937}                                                     
\newlength{\augPanelH}  \setlength{\augPanelH}{\dimexpr4\figRowH-\figGap\relax}   
\newlength{\augGap}     \setlength{\augGap}{\dimexpr3\figGap/2\relax}             
\newlength{\augLowY}    \setlength{\augLowY}{\augGap}                             
\newlength{\augHighY}   \setlength{\augHighY}{\dimexpr\augLowY+\augPanelH+\augGap\relax}
\newlength{\augFigH}    \setlength{\augFigH}{\dimexpr\augHighY+\augPanelH\relax}        
\newlength{\augNowX}    \setlength{\augNowX}{\nowFrac\columnwidth}                      
\newlength{\augAxisX}   \setlength{\augAxisX}{\dimexpr3\figGap\relax}                   
\newlength{\augAxisEnd} \setlength{\augAxisEnd}{\dimexpr\columnwidth-\figGap-\figGap/5\relax}
\begin{tikzpicture}[x=1mm, y=1mm, font=\scriptsize, inner sep=0, outer sep=0]
  \useasboundingbox (0,0) rectangle (\columnwidth,\augFigH);
  \def\now{\augNowX}
  \begin{scope}[shift={(0mm,\augHighY)}]
    \draw[arr, cGray] (\augAxisX,0) -- (\augAxisEnd,0);
    \node[lbl, anchor=north east, text=cGray] at (\augAxisEnd,-0.6) {time};
    \draw[thick, cGray] (\now,-\figGap/4) -- (\now,\augPanelH-\figGap/2);
    \node[lbl, below=0.6mm, text=cGray] at (\now,-\figGap/4) {$t_0$};
    \node[anchor=north west, font=\scriptsize] at (0,\augPanelH) {(a) planner input};
    \draw[very thick, cHuman] plot[smooth] coordinates
      {(8,5) (14,5.5) (20,7) (27,8.5) (\now,9) (41,8.5) (48,7) (55,5.5) (62,5) (70,5)};
    \node[lbl, right, text=cHuman] at ([xshift=\figGap/2]70,5) {demo};
    \draw[very thick, cSynth] plot coordinates
      {(\now,9) (31,9.6) (28,10.8) (25,10.4) (22,11.9) (19,13.2) (16,12.9) (13,14.1) (10,15)};
    \node[lbl, anchor=south, text=cSynth] at (25,15.5) {drifted history};
    \draw[cIntent, thick] (\now,3.8) -- (70,3.8);
    \foreach \bx in {\now,70} \draw[cIntent, thick] (\bx,3.8) -- (\bx,4.4);
    \node[lbl, anchor=north, text=cIntent] at (48,3.2) {chunk: clean demo};
  \end{scope}
  \begin{scope}[shift={(0mm,\augLowY)}]
    \draw[arr, cGray] (\augAxisX,0) -- (\augAxisEnd,0);
    \node[lbl, anchor=north east, text=cGray] at (\augAxisEnd,-0.6) {time};
    \draw[thick, cGray] (\now,-\figGap/4) -- (\now,\augPanelH-3\figGap);
    \node[lbl, below=0.6mm, text=cGray] at (\now,-\figGap/4) {$t$};
    \node[anchor=north west, font=\scriptsize] at (0,\augPanelH) {(b) tracker command};
    \foreach \x in {10,22,40,50,62,70} {\draw[cGray, densely dotted] (\x,4) -- (\x,12.5);}
    \node[lbl, anchor=south, text=cGray] at (50,12.7) {replan};
    \draw[very thick, cHuman] plot[smooth] coordinates
      {(8,5) (14,5.5) (20,7) (27,8.5) (\now,9) (41,8.5) (48,7) (55,5.5) (62,5) (70,5)};
    \node[lbl, right, text=cHuman] at ([xshift=\figGap/2]70,5) {demo};
    \draw[very thick, cSynth] plot coordinates
      {(10.0,5.09) (12.0,5.91) (14.0,6.83) (16.0,7.93) (18.0,9.13) (20.0,10.33) (22.0,11.46)};
    \draw[very thick, cSynth] plot coordinates
      {(22.0,7.46) (24.0,7.62) (26.0,7.72) (28.0,7.70) (30.0,7.59) (32.0,7.40) (34.0,7.13) (36.0,6.78) (38.0,6.35) (40.0,5.83)};
    \draw[very thick, cSynth] plot coordinates
      {(40.0,8.63) (42.0,8.84) (44.0,8.93) (46.0,8.96) (48.0,9.00) (50.0,9.04)};
    \draw[very thick, cSynth] plot coordinates
      {(50.0,6.54) (52.0,5.66) (54.0,4.83) (56.0,4.12) (58.0,3.52) (60.0,2.99) (62.0,2.50)};
    \draw[very thick, cSynth] plot coordinates
      {(62.0,5.00) (64.0,5.51) (66.0,6.07) (68.0,6.64) (70.0,7.20)};
    \foreach \x/\y in {34.00/7.13, 35.10/6.94, 36.20/6.74, 37.30/6.50, 38.40/6.25, 39.50/5.96} {\fill[cIntent] (\x,\y) circle (0.5);}
    \node[lbl, anchor=south, text=cSynth] at (16,13.2) {drifted command};
    \node[lbl, anchor=north west, text=cIntent] at ([xshift=1mm]\now,3.4) {command window};
  \end{scope}
\end{tikzpicture}%
  \caption{Augmentation of each policy's input, drawn for one pose coordinate over time. The demo curve is the demonstrated motion. (a)~Planner input at its observation time $t_0$. The drifted history is the demonstrated five-link history displaced by a random walk that starts at $t_0$ and runs back in time. The chunk, the planner's training target, stays the clean demonstration. (b)~Tracker command at its control step $t$. The drifted command is the demonstrated command plus a drift that restarts from zero at each replan (dotted lines), when the planner sends a new chunk. Blue dots mark the command window, the five-link command that the tracker reads at step $t$: six samples over the next \qty{0.2}{\second}, a short part of the \qty{1.16}{\second} chunk.}
  \label{fig:aug}
\end{figure}

\subsubsection{Planner Augmentation}
\label{sec:planner_drift}
At each history time, the drift augmentation moves the five links as one rigid body, by a rigid motion in the world frame.
Along the history, this motion follows a random walk: it is the identity at $t_0$ and accumulates random steps toward older times.
The steps are larger when the torso moves more over the history.
The motion also tilts the torso, so projected gravity and the torso height follow it.
Each wrist and each foot also follows a smaller random walk of its own.
The chunk stays clean (Fig.~\ref{fig:aug}(a)).

The camera is fixed to the torso, so the rigid motion also moves the camera.
The camera motion is the camera pose after the drift, expressed in the camera frame before the drift.
We warp each image of the history by the camera motion at its own time (Fig.~\ref{fig:imgaug}(c)).
To first order, the interaction matrix~\cite{chaumette2006visual} maps the camera motion to a 2D shift of each pixel.
This shift also depends on the metric depth of the pixel, which comes from~\S\ref{sec:planner_pixels}.
Each pixel of the warped image copies the original image at the pixel's position minus its shift.
The background of the images therefore moves consistently with the five-link history.

Each training sample also receives one random change of brightness, contrast, and lighting gradient, shared by its four images (Fig.~\ref{fig:imgaug}(b)).
We also mask samples and links of the five-link history at random, so the planner has to use the images.

\subsection{Whole-Body Tracker}
\label{sec:tracker}

The whole-body tracker turns the five-link command into joint setpoints.
It has to reach the commanded wrist poses and keep the robot balanced when the command drifts or a person pushes the robot.
It is a reinforcement-learning policy that builds on the motion tracker of BeyondMimic~\cite{liao2025beyondmimic}.

The five-link command holds six samples of the chunk, \qty{0.04}{\second} apart, from the current time $t$ to $t+\qty{0.2}{\second}$.
It is expressed in the robot's current torso frame, so the torso pose of its first sample is the current torso tracking error.
The observation is the current proprioception only: the angular velocity from the inertial measurement unit, the joint positions and velocities, and the previous joint setpoints.

\begin{table}[t]
  \caption{Reward kernels of the whole-body tracker.}
  \label{tab:reward}
  \centering
  \footnotesize
  \begin{tabular}{@{}lll@{}}
    \toprule
    Kernel & Frame & Width $\sigma$ \\
    \midrule
    \multicolumn{3}{@{}l}{\textit{Style}~\cite{liao2025beyondmimic}} \\
    Pelvis attitude & world & \qty{0.4}{\radian} \\
    Link pose, all 14 & pelvis $xy$, yaw & \qty{0.3}{\metre}, \qty{0.4}{\radian} \\
    Link velocity, all 14 & world & \qty{1}{\metre\per\second}, \qty{3.14}{\radian\per\second} \\
    \midrule
    \multicolumn{3}{@{}l}{\textit{Task}} \\
    Torso, wide & world & \qty{0.3}{\metre}, \qty{0.4}{\radian} \\
    Torso, narrow & world & \qty{0.06}{\metre}, \qty{0.15}{\radian} \\
    Wrists & world & \qty{0.06}{\metre}, \qty{0.15}{\radian} \\
    Feet & torso & \qty{0.06}{\metre}, \qty{0.1}{\radian} \\
    Swing-foot clearance, \qty{10}{\centi\metre} & height & \qty{0.035}{\metre} \\
    \bottomrule
  \end{tabular}
\end{table}

\subsubsection{Reward}
\label{sec:tracker_reward}
Five links do not determine the rest of the body, so the reward needs a whole-body pose.
Before training, an inverse-kinematics (IK) solve places the robot on the recorded five-link targets, with every stance foot flat on the floor.
This solve is plain IK and not retargeting: it does not scale the motion, and its solution need not be consistent over time.
The joints can even flip between two time steps.
We accept this because the IK pose serves only the style reward, the start state of each episode, and the image edit of~\S\ref{sec:planner_pixels}.

The reward is a sum of Gaussian kernels with unit weight (Table~\ref{tab:reward}), minus penalties on the rate of the joint setpoints and on joint-limit violations.
The style reward follows BeyondMimic: it scores all 14 links of the IK pose and keeps the motion natural.
The task reward scores the torso and the wrists in the world frame, so the wrists reach the commanded poses.
It scores the feet in the torso frame, so the robot can step where it needs to for balance.
A clearance kernel rewards \qty{10}{\centi\metre} of ankle height whenever the recorded foot swings.

\subsubsection{Command Augmentation}
\label{sec:tracker_aug}
At deployment, the command comes from a chunk, moved into the current torso frame with the onboard state estimate (\S\ref{sec:bridge}).
Until the planner replans, the command drifts with the errors of the state estimate, the planner, and the tracker.
The new chunk comes from a new image, so the command jumps back at the replan.
Training reproduces this drift and jump on the recorded trajectory (Fig.~\ref{fig:aug}(b)).
In training, a replan occurs at a random interval of \qtyrange{0.2}{1}{\second}.
Between two replans, one rigid drift of the five-link command grows linearly from zero.
It reaches at most \qty{10}{\centi\metre} of horizontal shift and \ang{10} of yaw about the recorded torso.
Each sample of the command carries the drift extrapolated to its own time.
We compute the reward against the drifted command, because the tracker cannot separate drift from intended motion.
The planner corrects the resulting error at its next replan.

\subsubsection{Training}
\label{sec:tracker_train}
Each episode starts from the IK pose with a random perturbation, at a demonstration time drawn by the adaptive sampler of BeyondMimic.
We randomize mass, friction, encoder bias, and sole geometry.
While the demonstrator holds a box in the demonstration, a spring replaces the box in simulation.
It can only push the wrists apart, with up to \qty{40}{\newton}.
For the suitcase task, the simulation contains a model of the suitcase, because the demonstrations hold no contact wrench to apply in its place.
Following SuperMimic~\cite{anonymous2026supermimic}, we optimize the whole-body tracker through a learned model of the simulator.
This model is a dynamics network fit to transitions simulated in mjlab~\cite{zakka2026mjlab}.
We train an actor and a critic on rollouts that this network predicts.
We train one tracker for each robot and each task, with no real-robot data.

\subsection{Deployment}
\label{sec:deploy}

\begin{figure}[t]
  \centering
\newlength{\dpAxisY}    \setlength{\dpAxisY}{\dimexpr4\figGap\relax}                    
\newlength{\dpCmdY}     \setlength{\dpCmdY}{\dimexpr\dpAxisY+\figRowH\relax}            
\newlength{\dpReadGap}  \setlength{\dpReadGap}{\dimexpr3\figGap/4\relax}                
\newlength{\dpImgY}     \setlength{\dpImgY}{\dimexpr\dpCmdY+\figRowH+\dpReadGap\relax}  
\newlength{\dpStateY}   \setlength{\dpStateY}{\dimexpr\dpImgY+\figRowH\relax}           
\newlength{\dpMidY}     \setlength{\dpMidY}{\dimexpr\dpImgY-3\figRowH/5\relax}          
\newlength{\dpTopY}     \setlength{\dpTopY}{\dimexpr\dpStateY+\figRowH-\figGap/4\relax} 
\newlength{\dpLblY}     \setlength{\dpLblY}{\dimexpr\dpAxisY-\figGap/4\relax}           
\newlength{\dpLblX}     \setlength{\dpLblX}{\dimexpr\figChan+\figGap\relax}             
\newlength{\dpHistX}    \setlength{\dpHistX}{\dimexpr\dpLblX+\figGap\relax}             
\newlength{\dpAxisX}    \setlength{\dpAxisX}{\dimexpr\dpLblX+\figGap/2\relax}           
\newlength{\dpAxisEnd}  \setlength{\dpAxisEnd}{\dimexpr\columnwidth-\figGap/5\relax}    
\newlength{\dpChunkEnd} \setlength{\dpChunkEnd}{\dimexpr\dpAxisEnd-\figGap\relax}       
\newlength{\dpTobs}     \setlength{\dpTobs}{0.3937\columnwidth}                         
\newlength{\dpTnow}     \setlength{\dpTnow}{0.53266\columnwidth}                        
\newlength{\dpOvEnd}    \setlength{\dpOvEnd}{\dimexpr\dpTnow+8\figGap\relax}            
\newlength{\dpDotR}     \setlength{\dpDotR}{\dimexpr\figTagGap-\figGap/4\relax}         
\newlength{\dpCmdDotR}  \setlength{\dpCmdDotR}{\dimexpr\figRowH/10\relax}               
\newlength{\dpHistStep} \setlength{\dpHistStep}{\dimexpr(\dpTobs-\dpHistX)/8\relax}     
\newlength{\dpImgStep}  \setlength{\dpImgStep}{\dimexpr2\dpHistStep\relax}              
\newlength{\dpFrameStep}\setlength{\dpFrameStep}{\dimexpr9\figGap/5\relax}              
\newlength{\dpFigH}     \setlength{\dpFigH}{\dimexpr\dpTopY+\figGap/2\relax}            
\begin{tikzpicture}[x=1mm, y=1mm, font=\scriptsize, inner sep=0, outer sep=0,
    rowlbl/.style={font=\scriptsize\bfseries, anchor=east}]
  \useasboundingbox (0,0) rectangle (\columnwidth,\dpFigH);
  \fill[cLight] (\dpTobs,\dpAxisY) rectangle (\dpTnow,\dpTopY);
  \node[lbl, anchor=north, text=cGray] at ([yshift=-\figGap/2]$(\dpTobs,\dpTopY)!0.5!(\dpTnow,\dpTopY)$) {camera\\delay +\\inference};
  \draw[arr, cGray] (\dpAxisX,\dpAxisY) -- (\dpAxisEnd,\dpAxisY);
  \node[lbl, anchor=north east, text=cGray] at (\dpAxisEnd,\dpLblY) {$t$};
  \draw[very thick, cGray] (\dpTobs,\dpAxisY) -- (\dpTobs,\dpTopY);
  \node[lbl, anchor=north east, align=right, text=cGray]
    at (\the\dimexpr\dpTobs+\figTagGap\relax,\dpLblY) {$t_0$\\image captured};
  \draw[very thick, cGray, dashed] (\dpTnow,\dpAxisY) -- (\dpTnow,\dpTopY);
  \node[lbl, anchor=north west, align=left]
    at (\the\dimexpr\dpTnow+\figGap/2\relax,\dpLblY) {chunk\\arrives};
  \node[rowlbl, text=cRobot] at (\dpLblX,\dpStateY) {five-link hist.};
  \draw[cRobot, thick] (\dpHistX,\dpStateY) -- (\dpTobs,\dpStateY);
  \foreach \i in {0,...,8}
    {\fill[cRobot] (\the\dimexpr\dpHistX+\i\dpHistStep\relax,\dpStateY) circle (\dpDotR);}
  \node[rowlbl, text=cRobot] at (\dpLblX,\dpImgY) {image hist.};
  \foreach \i in {0,1,2,3}
    {\draw[cRobot, thick, fill=white]
       (\the\dimexpr\dpTobs-\i\dpImgStep-\figTagGap\relax,\the\dimexpr\dpImgY-\figTagGap\relax)
       rectangle
       (\the\dimexpr\dpTobs-\i\dpImgStep+\figTagGap\relax,\the\dimexpr\dpImgY+\figTagGap\relax);}
  \node[rowlbl, text=cIntent] at (\dpLblX,\dpCmdY) {chunk};
  \fill[cIntent, opacity=0.15] (\dpTnow,\the\dimexpr\dpCmdY-\figRowH/3\relax)
    rectangle (\dpOvEnd,\the\dimexpr\dpCmdY+\figRowH/3\relax);
  \draw[cIntent, thick] (\dpTobs,\dpCmdY) -- (\dpChunkEnd,\dpCmdY);
  \foreach \i in {0,...,3}
    {\draw[cIntent, thick, fill=white] (\the\dimexpr\dpTobs+\i\dpFrameStep\relax,\dpCmdY) circle (\dpCmdDotR);}
  \foreach \i in {4,...,13}
    {\fill[cIntent] (\the\dimexpr\dpTobs+\i\dpFrameStep\relax,\dpCmdY) circle (\dpCmdDotR);}
  \node[lbl, anchor=north, text=cIntent]
    at (\the\dimexpr\dpOvEnd+5\figGap\relax,\the\dimexpr\dpCmdY-\figRowH/3-\figGap/4\relax)
    {action $k$ at $t_0+k\,\Delta t$};
  \draw[cGray, thick, dashed] (\dpTnow,\the\dimexpr\dpCmdY-\figGap/2\relax)
    -- (\dpOvEnd,\the\dimexpr\dpCmdY-\figGap/2\relax);
  \node[lbl, anchor=west, text=cGray] at (\the\dimexpr\dpTnow+\figGap\relax,\dpMidY)
    {overlap with previous chunk};
  \draw[arr, cRobot, very thick] (\dpTnow,\the\dimexpr\dpImgY-\figGap/2\relax)
    -- (\dpTnow,\the\dimexpr\dpCmdY+\figRowH/5\relax);
  \node[lbl, anchor=west, text=cRobot]
    at (\the\dimexpr\dpTnow+\figTagGap\relax,\dpImgY) {tracker reads at $t$};
\end{tikzpicture}%
  \caption{Timestamps at deployment. Both histories end at $t_0$, the capture time of the newest image. Hollow dots are actions that are already past when the chunk arrives.}
  \label{fig:deploy}
\end{figure}
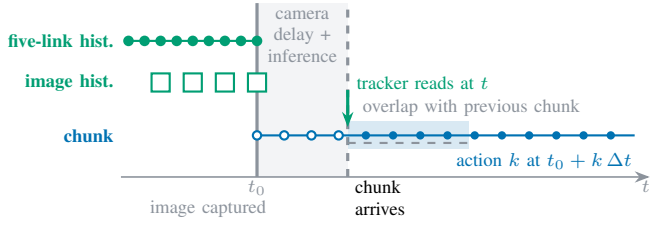

The planner replans at \qty{5}{\hertz}, and the tracker runs at \qty{50}{\hertz}.
The camera and the state estimator deliver their values with different delays, and the inference adds its own.
Following UMI~\cite{chi2024universal}, we measure each sensor delay once and stamp every value with the time that it describes (Fig.~\ref{fig:deploy}).
Action $k$ of a chunk, counted from zero, has the timestamp $t_0+k\Delta t$.
When a chunk arrives, its first actions are already in the past, and the tracker skips them.
At every control step $t$, the tracker interpolates the chunk at the six sample times of its command.
It uses linear interpolation in position and spherical linear interpolation in rotation.
It then moves the result into the torso frame of $t$ (\S\ref{sec:bridge}).

The planner infers consecutive chunks independently, so they can disagree where they overlap.
We move the overlapping actions of the previous chunk into the torso frame of the new $t_0$.
Following real-time chunking~\cite{black2025real}, we inpaint them into the new chunk during the flow integration, so the new chunk continues the previous one.
With the timestamps and the inpainted overlap, the planner can run slower than the tracker and deliver its chunks late.

Together, the shared interface, the image edit, the two augmentations, and the timestamps let two separately trained policies run as one system.

\section{SYSTEM EVALUATION}
\label{sec:eval}

On a real Unitree G1, three tasks demonstrate long-horizon, contact-rich, and reactive loco-manipulation, and box transport shows recovery from perturbations (\S\ref{sec:capability}).
In simulation, we ablate the design choices (\S\ref{sec:ablation}) and retrain both policies for a second humanoid from the same demonstrations (\S\ref{sec:crossemb}).

\begin{figure*}[t]
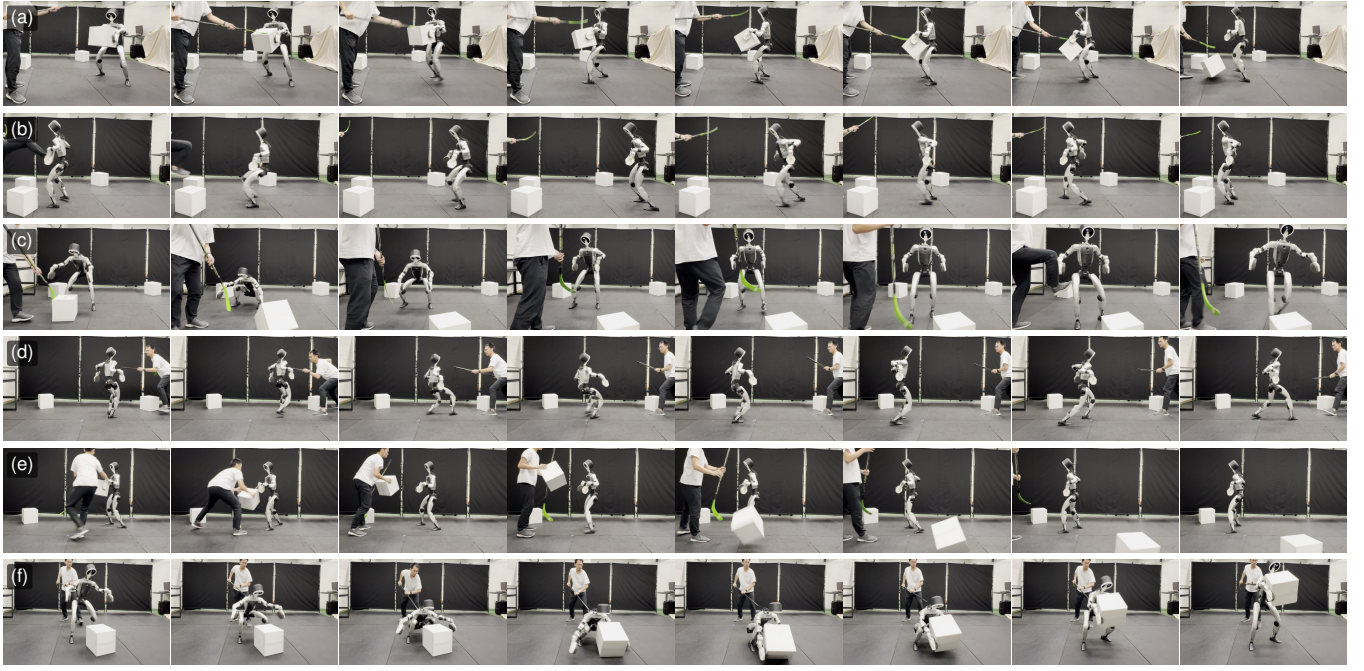

  \centering
  \perturbgrid{\textwidth}
  \caption{Recovery from perturbations during box transport on the real G1, in one continuous run. Each row shows eight frames at equal time intervals, from \qty{0.34}{\second} to \qty{0.61}{\second} depending on the row. A person applies each perturbation by hand, with a kick, or with a stick. (a)~Carried box pushed out of the arms. (b)~Kick to the torso; the robot takes several steps and walks to the nearest box. (c)~Box pulled away during the reach, then a kick; the robot stays on its feet and reaches again. (d)~Hard push on the torso while walking; the robot steps back and keeps walking. (e)~Box taken from the hands and thrown; the robot turns toward where it landed. (f)~Press on the back during the squat; the robot completes the lift.}
  \label{fig:results}
\end{figure*}

\subsection{Whole-Body Capability}
\label{sec:capability}

On the real Unitree G1, the whole-body tracker runs onboard and the planner runs on an external laptop.
The robot uses no external tracking, and the planner reads the five-link history from the robot's own proprioception.
The real-robot experiments demonstrate capability, and we report success rates only in simulation (\S\ref{sec:ablation}).
The three tasks share the interface, the training procedure, and the deployment software.
Each task has its own planner and tracker, both trained on that task's data.

\textit{Box transport.}
The robot walks to a box, squats, grasps it with both hands, and carries it to the shelf (Fig.~\ref{fig:teaser}(a)).
It places two boxes on the upper shelf this way, straightens them, and kicks the third box into the bottom shelf.
One episode handles the three boxes one after another without a reset.

A person pushes and kicks the robot during the task (Fig.~\ref{fig:results}).
In each kick and push shown, the whole-body tracker keeps the robot on its feet, and the planner continues the task.
When a person takes the box away, the planner sees this in the images, and the robot reaches again or walks to the nearest box.
The demonstrations contain perturbed objects but no push on the demonstrator (\S\ref{sec:protocol}), so none shows a recovery from a push.

\textit{Box catching.}
The robot catches a thrown box in the air with both hands and sets it down (Fig.~\ref{fig:teaser}(b)).
The thrower sets the timing and the trajectory, so the robot has to react to both.
The flight shown lasts about \qty{0.56}{\second}.

\textit{Suitcase pushing, toppling, and climbing.}
The robot pushes a \qty{14}{\kilo\gram} rolling suitcase to the shelf with its hands.
It topples the suitcase flat with its hands and one leg, which it presses against the base.
It steps up \qty{0.39}{\metre} onto the suitcase, takes a box from the top shelf, and jumps down (Fig.~\ref{fig:teaser}(c)).

\subsection{Ablation}
\label{sec:ablation}

\begin{figure}[t]
  \centering
  \panel[(a)]{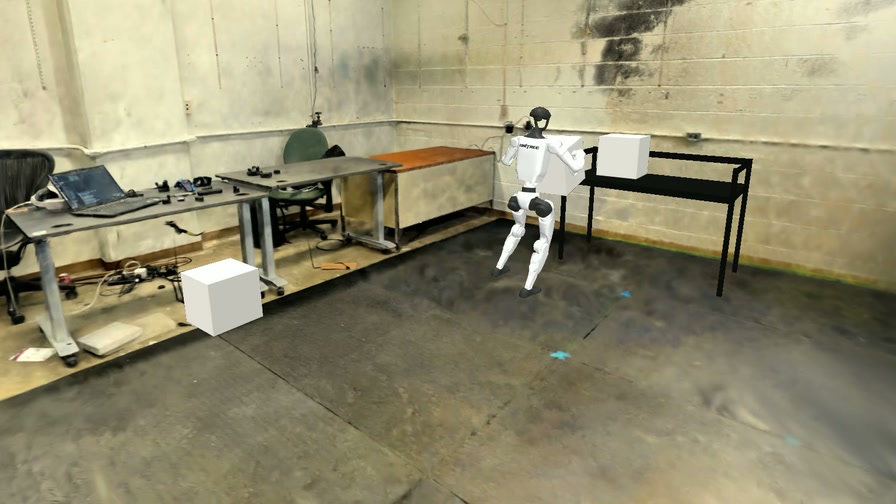}{\panelw{2}{\linewidth}}{G1}\hfill%
  \panel[(b)]{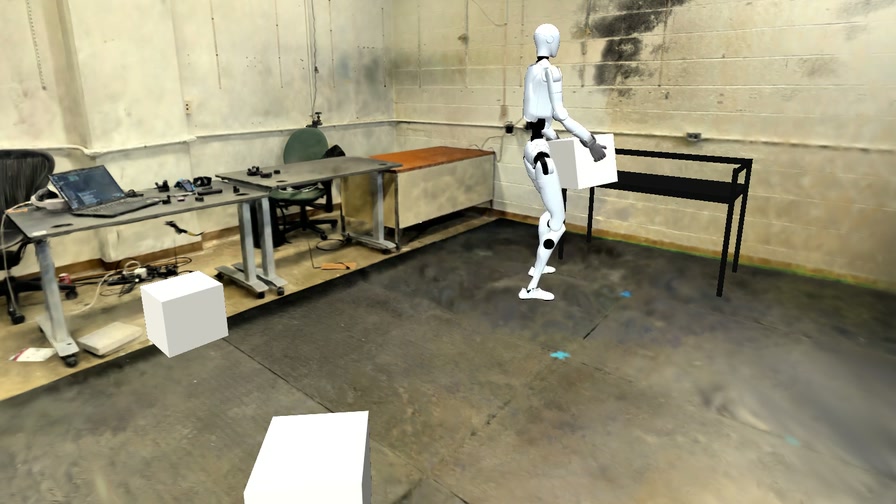}{\panelw{2}{\linewidth}}{H2}\\[\gut]
  \panel[(c)]{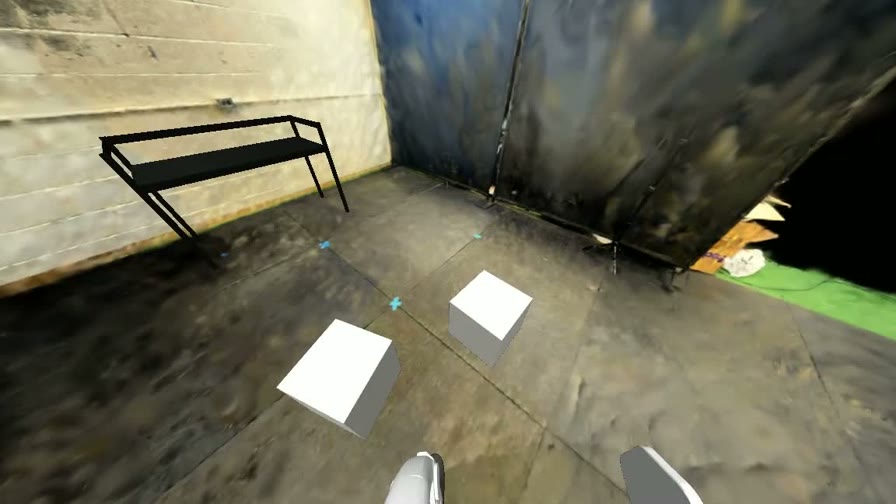}{\pthree}{approach}\hfill%
  \panel{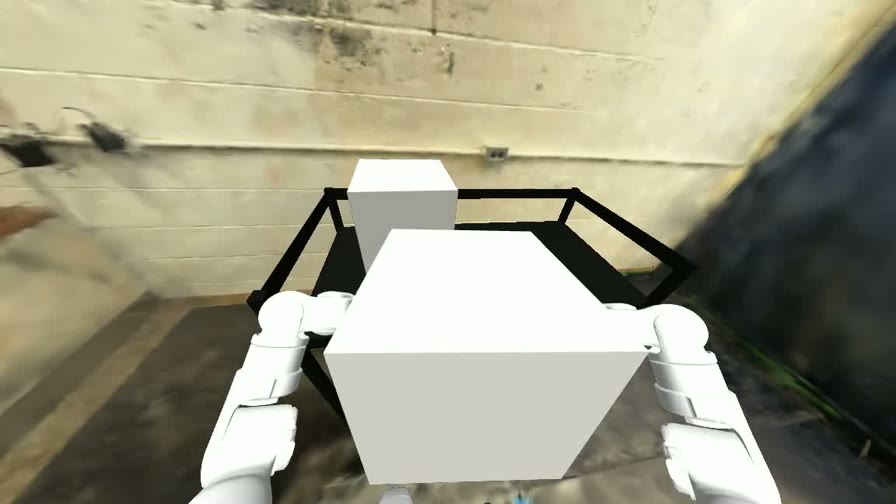}{\pthree}{carry}\hfill%
  \panel{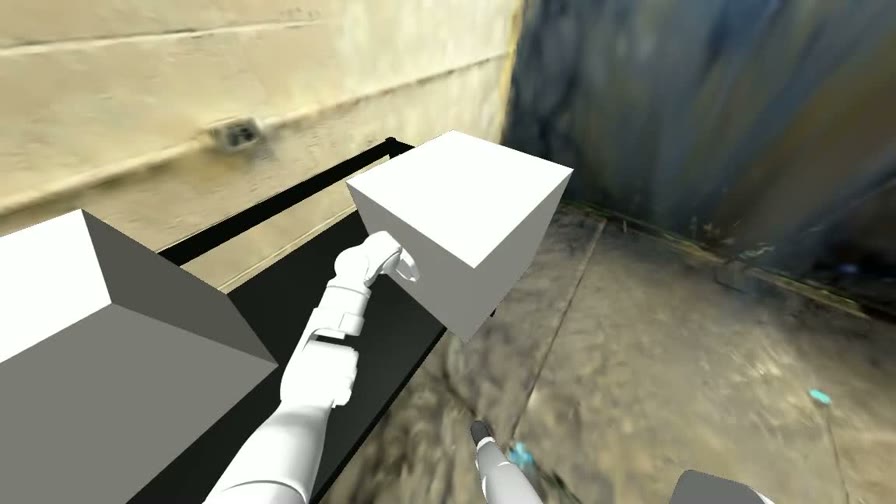}{\pthree}{place}
  \caption{Evaluation in the reconstructed room, a simulated copy of the demonstration room with a Gaussian-splat background. (a)~G1 and (b)~H2 on box transport. (c)~Egocentric images that the planner receives in simulation.}
  \label{fig:gs}
\end{figure}

\begin{table}[t]
  \centering
  \caption{Ablation on box transport in the reconstructed room.}
  \label{tab:ablation}
  \footnotesize\setlength{\tabcolsep}{2pt}
  \begin{tabular}{@{}lcccc@{\hspace{6pt}}cccc@{}}
    \toprule
    & \multicolumn{4}{c}{Without pushes} & \multicolumn{4}{c}{With pushes} \\
    \cmidrule(lr){2-5}\cmidrule(lr){6-9}
    & $\geq$1 & $\geq$2 & Success & Fall
    & $\geq$1 & $\geq$2 & Success & Fall \\
    \midrule
    \textbf{Ours (full system)} & 91 & \textbf{84} & \textbf{77} & \textbf{1} & 86 & \textbf{68} & \textbf{64} & 15 \\
    \midrule
    Joint command~\cite{wang2026bifrostumi} & 73 & 58 & 54 & \textbf{1} & 86 & 50 & 42 & \textbf{4} \\
    \midrule
    Link-local action frame   & 34 &  9 & 5  & 77 & 21 &  5 & 4  & 91 \\
    w/o projected gravity     & 24 &  5 & 1  & 97 & 13 &  2 & 1  & 99 \\
    \midrule
    w/o planner augmentation  & 36 & 25 & 20 & 79 &  7 &  1 & 0  & 95 \\
    w/o command augmentation  & \textbf{93} & 77 & 76 &  4 & \textbf{98} & 67 & 52 & 23 \\
    \bottomrule
  \end{tabular}

  \vspace{2pt}
  \parbox{\linewidth}{\footnotesize Entries are percentages of 100 simulated episodes. $\geq$1 and $\geq$2: at least one and at least two boxes placed. Success: all three boxes placed within \qty{120}{\second} with no fall. Bold marks the best value in each column.}
\end{table}

\textit{Reconstructed room.}
We run the simulated experiments in the reconstructed room, a simulated copy of the demonstration room (Fig.~\ref{fig:gs}(a)).
We scan the room with Polycam into a Gaussian splat~\cite{kerbl20233d} and render it in mjlab as the background.
Prior work joins such reconstructions to a physics engine to evaluate manipulation policies~\cite{li2024robogsim,zhang2025real} and to generate training data~\cite{li2024robogsim,kim2026legs}.
MuJoCo simulates the robot and the boxes, and we render them over the background with the intrinsics and resolution of the chest camera.
The planner therefore receives the same kind of egocentric image in simulation as on the real robot (Fig.~\ref{fig:gs}(c)).
A gap in appearance remains, and we reduce it by collecting about \qty{15}{\percent} of the planner's training data in the reconstructed room.
Only the simulated evaluation needs this part of the data.

\textit{Protocol.}
We compare the full system with five variants on box transport (Table~\ref{tab:ablation}).
All variants share the training data, the optimizer, and the schedule, and only the named component changes.
Each system runs 100 episodes without pushes and 100 with pushes.
Apart from the pushes, only the poses of the three boxes and the robot's start pose change between episodes.
Appearance, lighting, friction, and mass stay fixed.
A push is an impulse of \qty{40}{\newton\second} on the pelvis in a random direction: \qty{40}{\newton} for \qty{1}{\second}, every \qtyrange{5}{10}{\second}.
The full system succeeds in \qty{77}{\percent} of episodes without pushes and in \qty{64}{\percent} with pushes.
With 100 episodes, the \qty{95}{\percent} binomial interval of a rate near \qty{77}{\percent} is about $\pm8$ points.

\textit{Joint command.}
This variant gives the whole-body tracker a joint command in place of the five-link command.
An online inverse-kinematics solver computes the 29 joint positions from the five-link targets, as BifrostUMI~\cite{wang2026bifrostumi} does with its retargeting module.
Success drops to \qty{54}{\percent} without pushes and to \qty{42}{\percent} with pushes.
With pushes, the robot falls in \qty{4}{\percent} of episodes, against \qty{15}{\percent} for the full system.
A joint command is always reachable, which may explain the fewer falls.
The online solver is imperfect and can hit a singularity, which may explain the lower success.

\textit{Link-local action frame.}
This variant expresses the target of each link relative to the current pose of the same link.
Success drops to \qty{5}{\percent} without pushes and to \qty{4}{\percent} with pushes, and the robot falls in most episodes.
In the shared torso frame of the full system, the five poses show the body shape directly.
The planner then likely learns the targets more easily and sends fewer out-of-distribution chunks to the whole-body tracker.

\textit{Without projected gravity.}
This variant removes projected gravity and the torso height from the five-link history (\S\ref{sec:planner_io}).
Success drops to \qty{1}{\percent}, and the robot falls in almost every episode.
The planner then has to estimate the torso's tilt and height from the images.

\textit{Without planner augmentation.}
This variant removes all planner augmentation of~\S\ref{sec:planner_drift}, so it does not separate the effects of its parts.
Success drops to \qty{20}{\percent} without pushes and to \qty{0}{\percent} with pushes, and the robot falls in most episodes.
The drift augmentation likely teaches the planner to recover when the whole-body tracker does not follow a chunk exactly.

\textit{Without command augmentation.}
This variant removes the command augmentation of the whole-body tracker (\S\ref{sec:tracker_aug}).
Without pushes, success is \qty{76}{\percent}, against \qty{77}{\percent} for the full system.
With pushes, success drops from \qty{64}{\percent} to \qty{52}{\percent}, and falls rise from \qty{15}{\percent} to \qty{23}{\percent}.
The command augmentation therefore appears to matter mainly with pushes.
A push leaves the chunk far from the robot's state.
A whole-body tracker trained only on clean commands then receives a command outside its training distribution.

\subsection{Cross-Embodiment}
\label{sec:crossemb}

In simulation, we retrain both policies for the Unitree H2 from the same demonstrations, with no new recording (\S\ref{sec:align}, Fig.~\ref{fig:gs}(b)).
The H2 planner trains on images rendered in the reconstructed room, while the simulated H2 replays the demonstrations.
The H2 is a full-size humanoid, closer to the demonstrator in size and proportion than the G1 (Fig.~\ref{fig:bodies}).
Without pushes, the H2 completes box transport in \qty{83}{\percent} of 100 episodes, and the G1 in \qty{77}{\percent}.
The difference is smaller than the \qty{95}{\percent} interval of a difference between two such rates, about $\pm11$ points.

\subsection{Summary}

On the real G1, \projname{} performs the three tasks and recovers from the perturbations of Fig.~\ref{fig:results}.
In simulation, no variant succeeds more often than the full system, and the joint-command variant falls less often with pushes.
With the link-local action frame, or without projected gravity or planner augmentation, the robot falls in most episodes.
The command augmentation appears to matter mainly with pushes.
Without the augmentations, errors that the policies did not see in training likely grow over the long task and with pushes.
The same demonstrations also train both policies for a second humanoid in simulation.

\section{LIMITATIONS AND FUTURE WORK}

\textit{Scope of the results.}
Each task has its own planner and its own whole-body tracker.
One demonstrator recorded all the data in one room.
All tests take place in that room or in the reconstructed room.
The real-robot experiments demonstrate capability.
Success rates come from simulation only.
Under \qty{40}{\newton\second} pushes, the full system falls in \qty{15}{\percent} of the simulated episodes.

\textit{Real-world deployment on the H2.}
The H2 result is in simulation (\S\ref{sec:crossemb}).
Running the retrained policies on the real H2 is the next step.

\textit{A richer interface.}
The five links do not describe the fingers or the head.
Future work could add the state of a gripper or a dexterous hand to the interface.
It could also add a neck that carries the camera, so the robot can choose where it looks.

\textit{Recording outside one room.}
The Lighthouse base stations limit data collection to one room.
A pose estimate without base stations would let us record in other places and test the robot there.

\section*{ACKNOWLEDGMENT}
We thank Bike Zhang, Junfeng Long, Yiyang Shao, Chuye Hong, and Yen-Jen Wang for discussions and for their help with the experiments.
This work was supported in part by Komatsu and Apple.
We thank Lambda for sponsoring compute for this project.
Pieter Abbeel and Koushil Sreenath hold concurrent appointments as Professors at UC Berkeley and as Amazon Scholars.
This paper describes work performed at UC Berkeley and is not associated with Amazon.

\IEEEtriggercmd{\balance}\IEEEtriggeratref{17}  
\bibliographystyle{IEEEtran}
\bibliography{references}

\end{document}